\documentclass[conference]{IEEEtran}
\IEEEoverridecommandlockouts
\usepackage{cite}
\usepackage{amsmath,amssymb,amsfonts}
\usepackage{algorithmic}
\usepackage{graphicx}
\usepackage{textcomp}
\usepackage{xcolor}
\def\BibTeX{{\rm B\kern-.05em{\sc i\kern-.025em b}\kern-.08em
    T\kern-.1667em\lower.7ex\hbox{E}\kern-.125emX}}
\usepackage[hidelinks]{hyperref}
\usepackage{textcomp}
\usepackage{siunitx}
\usepackage{tabularx}
\usepackage[export]{adjustbox}
\usepackage{placeins}
\usepackage{nicematrix,enumitem}
\usepackage{booktabs}
\usepackage{ragged2e}
\usepackage{tikz}
\usepackage{graphicx}
\usetikzlibrary{arrows.meta, positioning, calc}
\usepackage{pgfplots}

\pgfplotsset{compat=1.9}
\usepgfplotslibrary{statistics}

\begin{document}

\bstctlcite{IEEEexample:BSTcontrol}

\title{Deep Learning for Generating Computational PIN-4 Immunohistochemistry Staining from Prostate Biopsy H\&E Images
\\
{\footnotesize \textsuperscript{}}
\thanks{}
}
\author{
\IEEEauthorblockN{Vietbao Tran}
\IEEEauthorblockA{\textit{Biomedical Engineering} \\
\textit{University of California, Irvine}\\
Irvine, CA, USA \\
vietbaot@uci.edu}
\and
\IEEEauthorblockN{Pratik Shah*\thanks{*Senior and corresponding author: Dr.\ Pratik Shah Ph.D. (pratik.shah@uci.edu)}}
\IEEEauthorblockA{\textit{Pathology and Laboratory Medicine} \\
\textit{Biomedical Engineering} \\
\textit{Electrical Engineering and Computer Science} \\
\textit{University of California, Irvine}\\
Irvine, CA, USA \\
pratik.shah@uci.edu}
}
\maketitle

\begin{abstract}
Immunohistochemistry (IHC) is frequently used to resolve diagnostically ambiguous prostate cancer biopsy findings on hematoxylin and eosin (H\&E)-stained tissue. However, PIN-4 IHC is typically performed on adjacent tissue sections, limiting direct spatial comparison between the H\&E morphology that indicated the need for IHC and the corresponding immunophenotypic signal. A paired, registered H\&E/PIN-4 dataset was constructed from routine clinical prostate biopsy whole-slide images (WSIs), and a conditional generative adversarial network (cGAN) was trained to synthesize PIN-4 staining patterns directly from native H\&E image patches. The final dataset comprised 172 paired WSIs from 93 patients and 27,298 registered $1024 \times 1024$ patch pairs, spanning adenocarcinoma-positive and benign cases with representation across age, race, and ethnicity groups. The model was evaluated on an independent held-out test set of 1,814 patch pairs from 17 WSIs, achieving a mean peak signal-to-noise ratio (PSNR) of 21.88\,dB, structural similarity index measure (SSIM) of 0.667, Pearson correlation coefficient (PCC) of 0.684, and learned perceptual image patch similarity (LPIPS) of 0.417. Qualitative review by a board-certified pathologist demonstrated that generated images captured diagnostically relevant PIN-4 staining patterns, including AMACR/racemase expression and basal-cell-associated staining, while preserving spatial correspondence with the source H\&E morphology. Synthesis quality varied across morphologically complex regions, including high-grade carcinoma and intraductal carcinoma. These results support the feasibility of supervised PIN-4 synthesis from routinely acquired brightfield H\&E prostate biopsy images. The approach enables direct interpretation of predicted PIN-4 marker patterns in the context of the source prostate H\&E architecture, addressing a current spatial limitation of conventional adjacent-section IHC.

\end{abstract}

\begin{IEEEkeywords}
Image-to-image translation, Prostate cancer, Deep learning, Computational staining
\end{IEEEkeywords}

\section{Introduction}

Prostate cancer is among the most diagnosed malignancies in men in the United States, with more than 333,000 new cases estimated annually \cite{siegel_cancer_2026}. It remains the second leading cause of cancer-related mortality among men, with more than 36,000 deaths estimated annually \cite{siegel_cancer_2026}. Accurate diagnosis and risk stratification rely on histopathologic evaluation of prostate tissue, most commonly from needle core biopsies.

\subsection{Histopathology Dye Staining and Cancer Diagnosis}

The standard diagnostic workflow for prostate cancer relies on brightfield microscopic examination of chemically stained hematoxylin and eosin (H\&E) tissue sections digitized as whole-slide images (WSIs). Pathologists assess glandular architecture and cytologic features on H\&E WSI and assign Gleason patterns to characterize tumor differentiation \cite{humphrey_histopathology_2017}. The primary and secondary Gleason patterns are combined to produce a Gleason score, with higher scores generally associated with more aggressive disease and increased prostate cancer mortality \cite{delahunt_gleason_2012}. In diagnostically ambiguous cases, immunohistochemistry (IHC) staining is often used to support differential diagnosis \cite{humphrey_histopathology_2017}. PIN-4 is a commonly used prostate IHC cocktail that combines basal cell markers, including p63 and high-molecular-weight cytokeratin (CK5/6), with alpha-methylacyl-CoA racemase (AMACR/P504S)\cite{sabata_automated_2010}. This combination supports differential diagnosis among benign glands, high-grade prostatic intraepithelial neoplasia (HGPIN), atypical small acinar proliferation (ASAP), and prostatic adenocarcinoma \cite{sabata_automated_2010}. However, currently PIN-4 IHC staining is performed on an adjacent tissue section rather than the original H\&E section. As a result, spatial loss, sectioning depth, deformation, and tissue architectural differences limit direct cell-level comparison between the H\&E morphology that prompted IHC ordering and the corresponding immunophenotypic signal on the adjacent section\cite{ravery2008twenty}

\subsection{ Computational Staining and Image Registration}
Computational staining uses deep learning-based image-to-image translation to transform pathology images between visual domains. Conditional generative adversarial neural networks (cGANs) have been applied to generate H\&E-like images from unstained tissue and to destain chemically stained WSIs \cite{ rana_use_2020}, enabling pathologist review while reducing chemical processing steps. Computational staining has also been extended across imaging modalities, including translation of photoacoustic microscopy images to H\&E-like outputs \cite{martell_deep_2023}, noninvasive \textit{in vivo} skin imaging for biopsy-free assessment \cite{li_biopsy-free_2021}, and autofluorescence-to-H\&E synthesis in autopsy tissue \cite{li_virtual_autopsy_2024}. Spatial alignment between H\&E and IHC WSIs is necessary for paired computational staining, patch-level learning, and direct morphologic comparison. Registration methods for histopathology images are commonly divided into global and deformable approaches \cite{chen_image_2023}. 

Global methods estimate a single transformation across the image, such as rigid, similarity, affine, or homography transformations. A homography uses a $3 \times 3$ matrix with eight degrees of freedom and can be estimated from matched image features, such as scale-invariant feature transform (SIFT) keypoints, followed by outlier rejection and transformation fitting \cite{faust_integrating_2022, shafique_automatic_2021, lippolis_automatic_2013}. Although global alignment can correct large-scale translation, rotation, scaling, and shearing, paired histologic sections often contain local tissue deformation caused by sectioning, staining, mounting, and scanning. Deformable registration methods address these local differences by estimating a spatially varying displacement field, typically constrained to preserve smoothness and continuity \cite{chen_image_2023}. Parametric approaches represent the deformation using basis functions such as B-splines, whereas nonparametric approaches iteratively optimize local displacement fields under regularization constraints \cite{chen_image_2023, klein_elastix_2010}. 
\subsection{Related Work}

Curent histopathology registration workflows often use global alignment for coarse initialization followed by deformable refinement to improve local correspondence between H\&E and IHC tissue structures \cite{chen_image_2023}. Recent computational staining work has also explored generating IHC marker patterns directly from H\&E images.  An unpaired GAN with a pathology-guided loss demonstrated multi-marker synthesis of glypican-3, human epidermal growth factor receptor 2 (HER2), and estrogen receptor (ER) across hepatocellular carcinoma and breast cancer datasets \cite{li_virtual_2024}. A paired cGAN incorporating task-specific membrane enhancement achieved superior perceptual quality for H\&E-to-HER2 translation, though at the cost of lower SSIM \cite{peng_advancing_2024}. For prostate cancer, a Pix2Pix-based model converted hyperspectral autofluorescence microscopy images into H\&E and PIN-4 outputs with Gleason grading agreement, though PIN-4 noninferiority was not demonstrated \cite{wong_clinical-grade_2024}. Existing prostate PIN-4 staining models have relied on specialized imaging inputs rather than standard brightfield H\&E~\cite{wong_clinical-grade_2024}, and current H\&E-to-IHC approaches have used unpaired frameworks without spatially registered H\&E/IHC image pairs~\cite{li_virtual_2024}. The present work addresses both limitations.

\subsection{Summary of Contributions}

In this study, we report a supervised H\&E-to-PIN-4 computational staining framework built from routine clinical prostate biopsy and resection WSIs collected at the University of California, Irvine Medical Center (UCI Health) in Orange County, CA. The final paired, registered dataset included 172 tissue blocks from 93 patients, 722 registered cores, and 27,298 paired $1024 \times 1024$ image patches after quality filtering, spanning adenocarcinoma-positive and benign cases with representation across age, race, and ethnicity groups. Model performance was evaluated on a held-out test set of 1,814 patch pairs from 17 tissue blocks using peak signal-to-noise ratio (PSNR), structural similarity index measure (SSIM), Pearson correlation coefficient (PCC), learned perceptual image patch similarity (LPIPS), and qualitative pathologist review. The contributions of this study are:

\begin{itemize}
    \item Development of a core-level registration and quality-control pipeline for aligning prostate biopsy WSIs before patch extraction, including review and exclusion of misregistered or artifact-containing regions.
    \item Demonstration that standard brightfield H\&E prostate biopsy WSIs contain sufficient learnable structure to support supervised PIN-4 immunophenotypic pattern synthesis without specialized inputs such as hyperspectral or autofluorescence microscopy.
    \item Training and quantitative evaluation of a supervised cGAN for synthesizing PIN-4 staining patterns from $1024 \times 1024$ brightfield H\&E patches, assessed on a held-out test set using PSNR, SSIM, PCC, and LPIPS.
    \item Qualitative pathologist review of computationally stained PIN-4 WSIs to assess preservation of diagnostically relevant staining relationships, providing clinical context for the quantitative metrics.
\end{itemize}

\section{Methods}

\subsection{Dataset}
Deidentified patient data and WSIs were acquired from UCI Health (IRB approval \#5733). The source cohort included 5,237 prostate core biopsies or resections from 4,155 patients. From this cohort, tissue blocks from 153 patients provided 394 H\&E and 233 PIN-4 glass slides, which were digitized at $40\times$ magnification using a Roche Ventana DP 600 scanner (Roche Diagnostics, Santa Clara, CA). After excluding unavailable or unpaired images and applying preprocessing, registration, quality-control, and patch-filtering steps, the final dataset included tissue blocks from 93 patients, comprising 172 H\&E/PIN-4 WSIs pairs and 27,298 registered patch pairs. Within the final cohort, 54 patients contributed adenocarcinoma-positive material, and 39 patients had benign-only biopsy results.

\subsection{Core Extraction from Whole-Slide Images}
Whole-slide images were loaded using TIAToolbox version 1.6.0 \cite{pocock_tiatoolbox_2022}. A modified version of \texttt{roi\_picker} was used to display WSIs at reduced resolution and support manual annotation of individual tissue cores as closed polygonal regions \cite{quy_mint-labroi_picker_2025}. For each H\&E WSI, a trained researcher delineated distinct tissue cores, and the resulting annotations were exported as JSON files. A Python extraction script then used these JSON annotations to extract each annotated WSI region within its rectangular bounding box at the second pyramidal level, corresponding to $20\times$ objective magnification. Extracted H\&E core images were saved losslessly as PNG files while retaining WSI metadata.

The same workflow was applied to PIN-4 WSIs, with the extraction script modified to load the corresponding H\&E core regions as initialization. The researcher then manually adjusted the PIN-4 polygons to align with corresponding H\&E serial-section cores while preserving shared region identifiers when possible. The H\&E annotation regions were removed when the corresponding core was absent in the PIN-4 WSI and added when unique PIN-4 cores were present. Exported core images were organized by stain type and tissue block. In total, 2,010 H\&E cores and 1,542 PIN-4 cores were extracted. The lower PIN-4 core count reflects the selective clinical ordering of PIN-4 IHC, cases where PIN-4 sections were unavailable, and exclusions due to staining or tissue artifacts.
\subsection{Registration}
Before registration, pixels outside each annotated core polygon were replaced with white to reduce background variability and focus alignment on the extracted tissue region. A modified version of VALIS version 1.2.0 \cite{gatenbee_valis_2025}, an open-source registration pipeline for large pathology images that performs sequential rigid and non-rigid alignment to correct serial-section tissue deformation, was tailored for path handling. In this study, VALIS used the DIScrete Keypoints (DISK) feature detector, LightGlue feature matcher, and random sample consensus (RANSAC) filtering for rigid alignment, followed by dense optical flow for non-rigid alignment. The registration workflow included initial rigid alignment, high-resolution rigid micro-registration, initial non-rigid registration, and high-resolution non-rigid micro-registration. Two modifications were made to the default VALIS configuration to optimize performance for prostate biopsy cores. First, the non-rigid micro-registration step was configured to downsample cores to 25\% resolution rather than the default maximum of 2048 pixels along the longest axis, improving deformable alignment for the elongated geometry of biopsy cores. Second, the registration background color was set to white to match the masked core background and prevent background contrast from interfering with feature detection. These adaptations enabled successful alignment of 974 paired cores from 207 paired WSIs.

To ensure registration quality, a trained researcher performed systematic visual review of all registered H\&E and PIN-4 core overlays in Sedeen Viewer version 5.4.4 \cite{noauthor_sedeen_nodate}. Closed polygonal exclusion regions were manually drawn to precisely delineate areas of tissue discontinuity, artifacts, poor image quality, or local registration error. These exclusion regions were exported as XML files and applied during downstream quality assessment and patch extraction to exclude the misaligned patch pairs from the training dataset.

\subsection{Registration Quality Metrics}
The Dice index between paired tissue masks was used as the primary objective registration quality metric \cite{zou_statistical_2004}. Normalized PCC, PSNR, and SSIM were also calculated on grayscale-converted core pairs as secondary registration metrics. Image quality and registration quality were additionally evaluated using five-point visual scoring rubrics. For image quality, scores of 1 or 2 indicated substantial blur that made the image unusable, whereas a score of 3 indicated blur, or artifact sufficient to exclude the image. Scores of 4 and 5 indicated usable images, with 4 representing good sharpness with and 5 representing near-perfect sharpness without relevant artifact. Registration quality was scored similarly, where scores of 1 or 2 indicated complete or severe misalignment, score 3 indicated substantial internal mismatch despite partial boundary alignment, score 4 indicated good alignment with minor local geometric differences, and score 5 indicated near-perfect alignment. Manually defined exclusion regions were omitted from image and registration quality assessment unless they covered most of the tissue region. Representative examples of registration quality scores are shown in Fig.~\ref{fig:qualitative_image_quality_examples}. Core pairs with both image quality and registration quality scores of 4 or 5 were retained for patch extraction, yielding 722 high-quality registered core pairs.

\begingroup
\setlength{\tabcolsep}{3pt}
\begin{figure}[htbp]
\centering
\begin{tabular}{l c c c c c}
\tiny{A} & \adjustbox{valign=c}{\includegraphics[width=0.15\linewidth,frame]{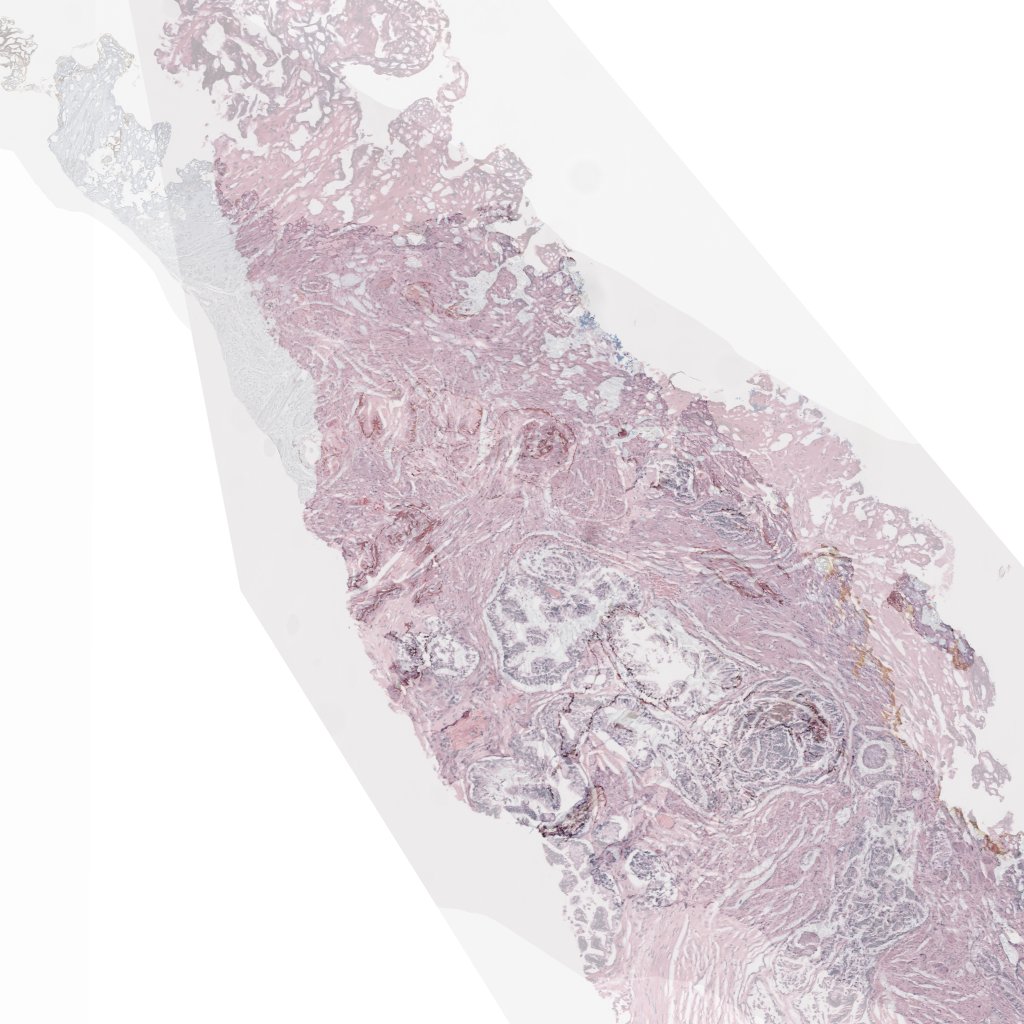}} & \adjustbox{valign=c}{\includegraphics[width=0.15\linewidth,frame]{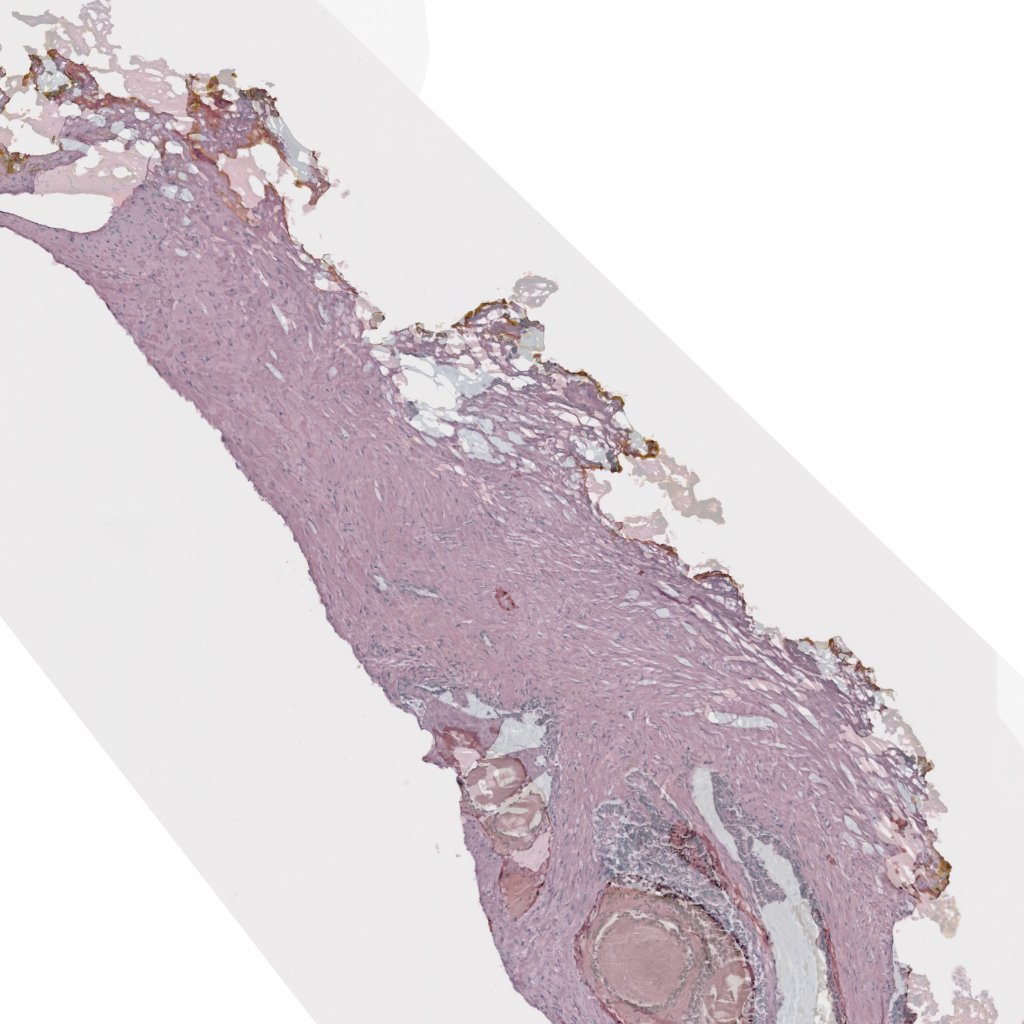}} & \adjustbox{valign=c}{\includegraphics[width=0.15\linewidth,frame]{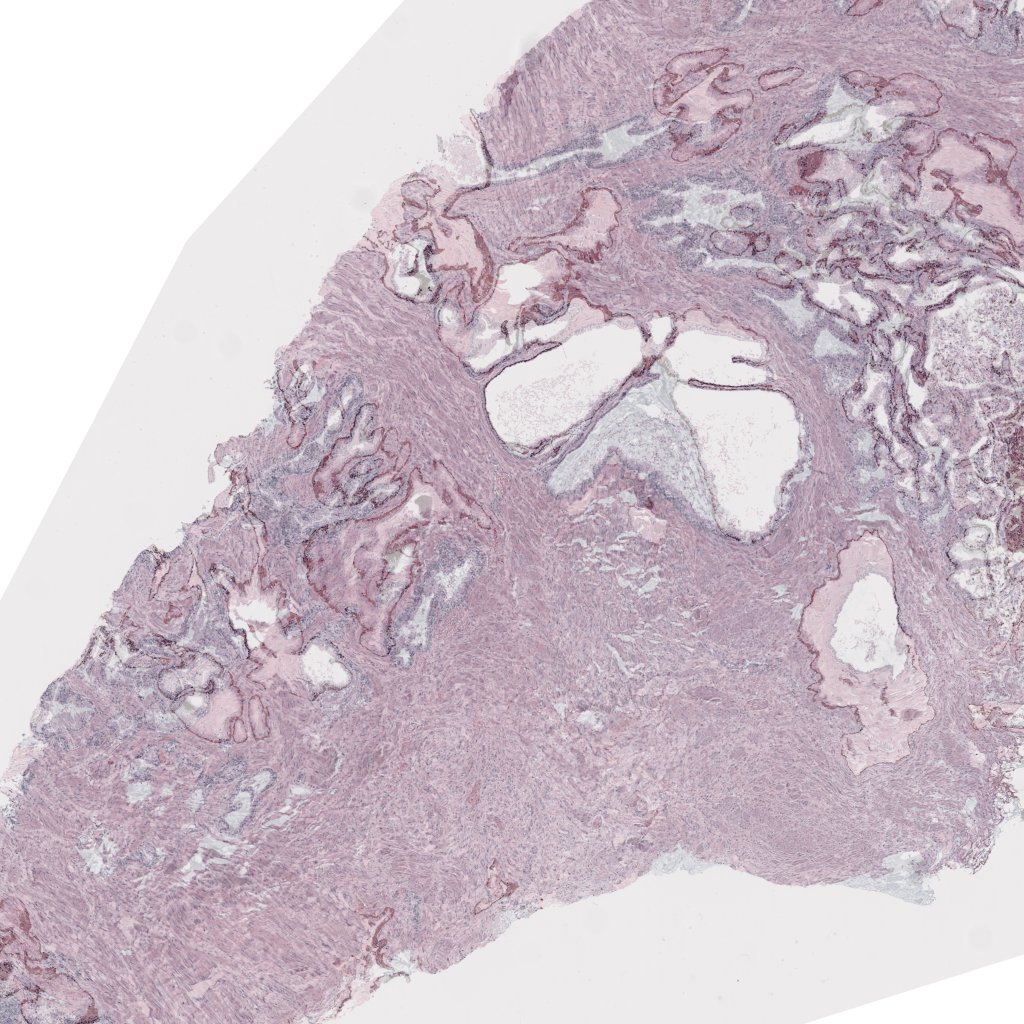}} & \adjustbox{valign=c}{\includegraphics[width=0.15\linewidth,frame]{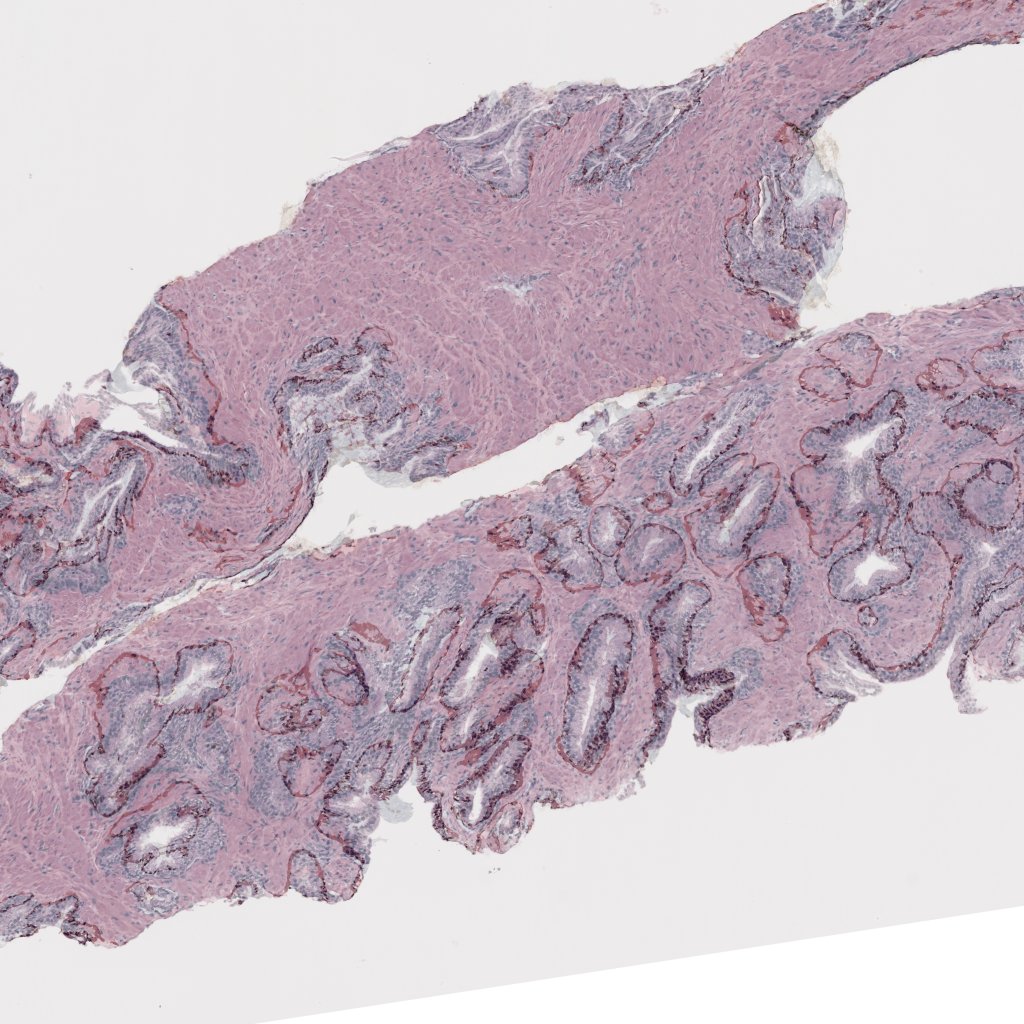}} & \adjustbox{valign=c}{\includegraphics[width=0.15\linewidth,frame]{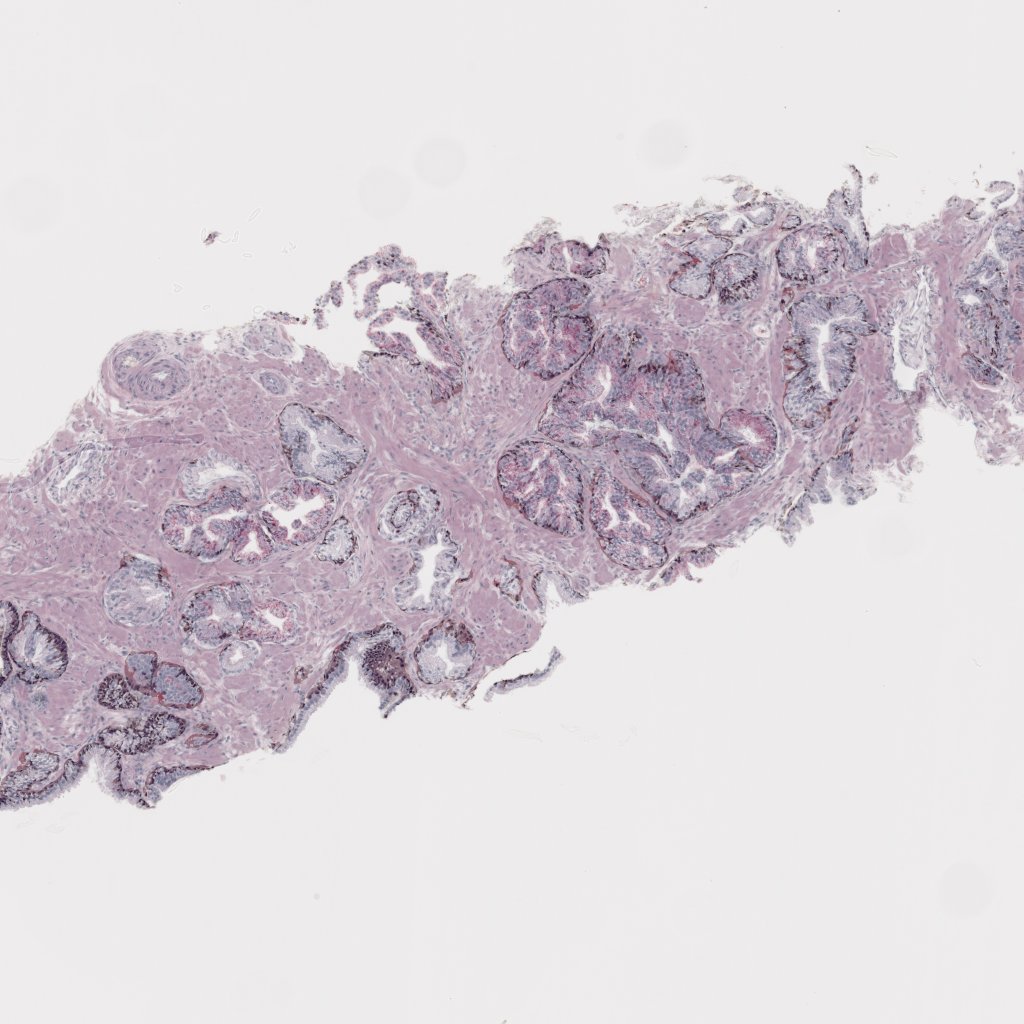}} \\[0.075\linewidth]
\tiny{B} & \adjustbox{valign=c}{\includegraphics[width=0.15\linewidth,frame]{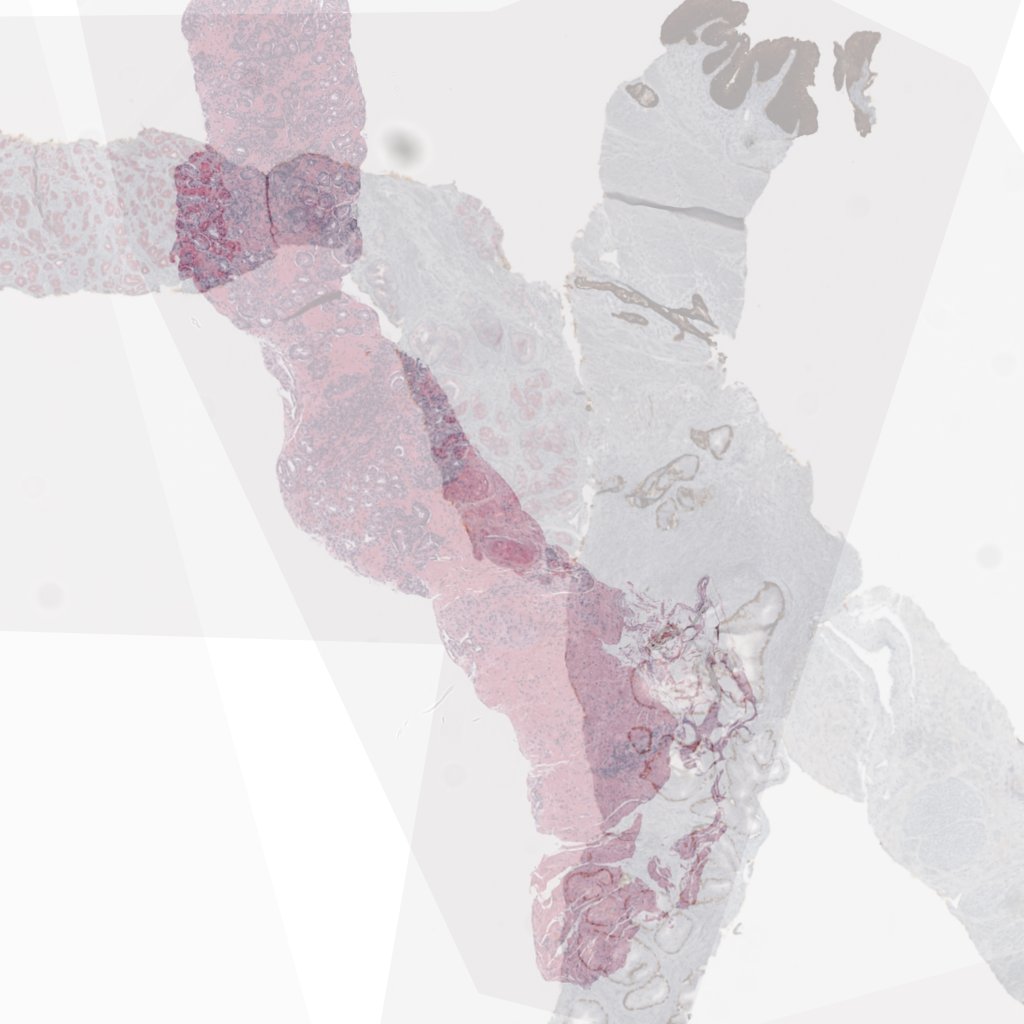}} & \adjustbox{valign=c}{\includegraphics[width=0.15\linewidth,frame]{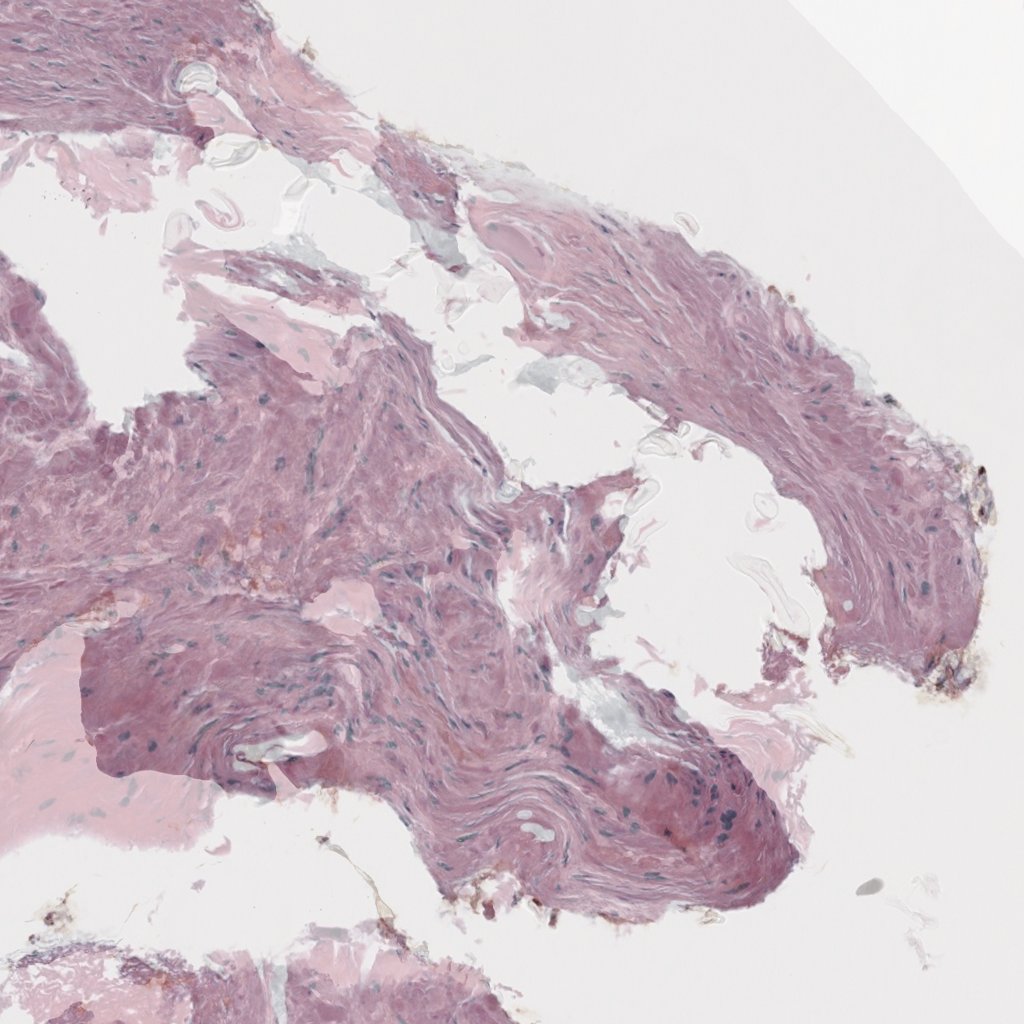}} & \adjustbox{valign=c}{\includegraphics[width=0.15\linewidth,frame]{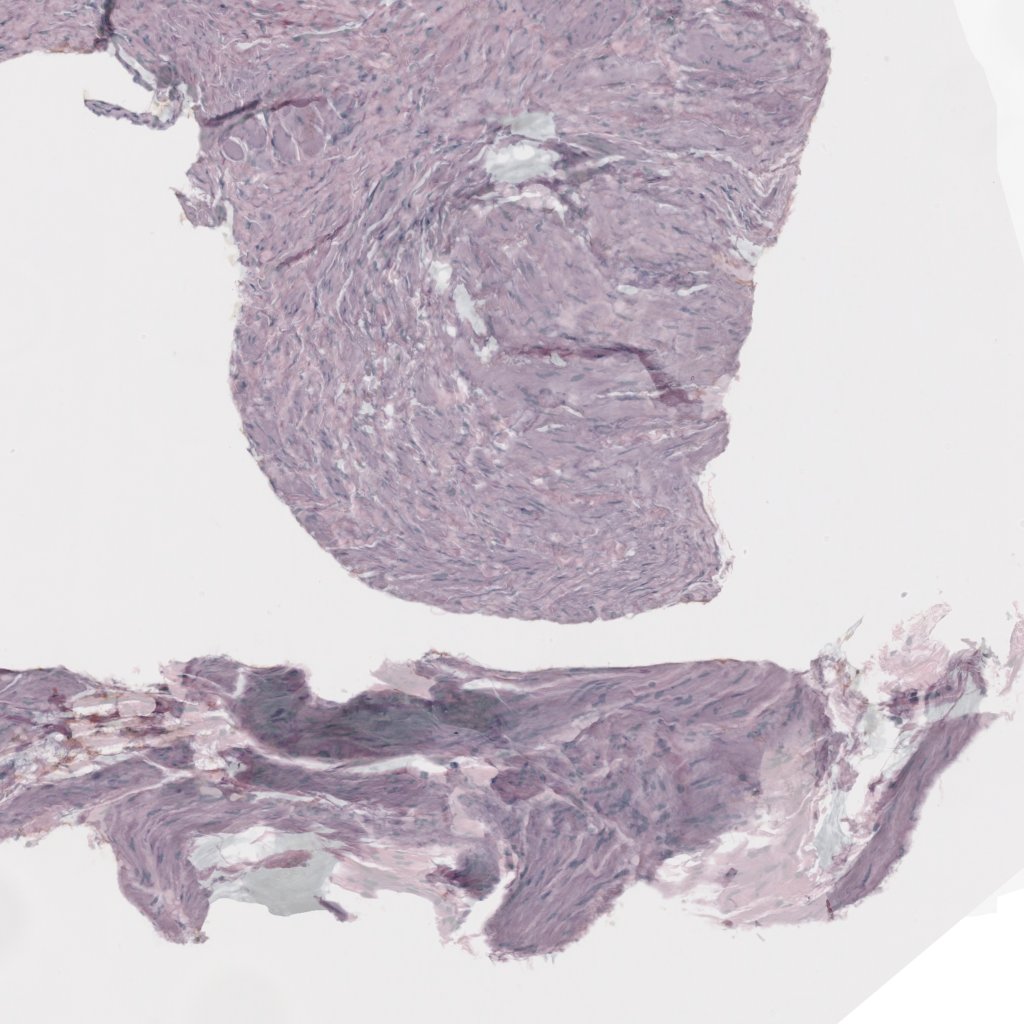}} & \adjustbox{valign=c}{\includegraphics[width=0.15\linewidth,frame]{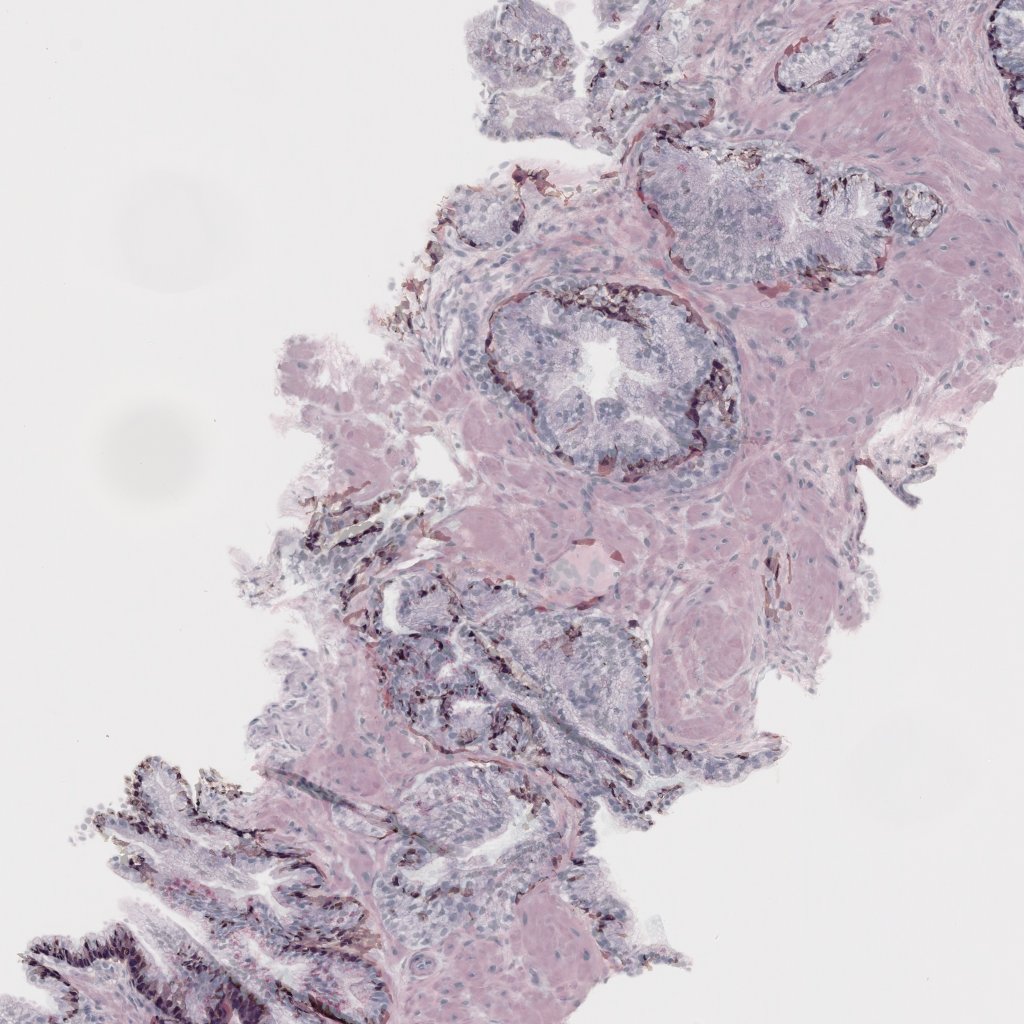}} & \adjustbox{valign=c}{\includegraphics[width=0.15\linewidth,frame]{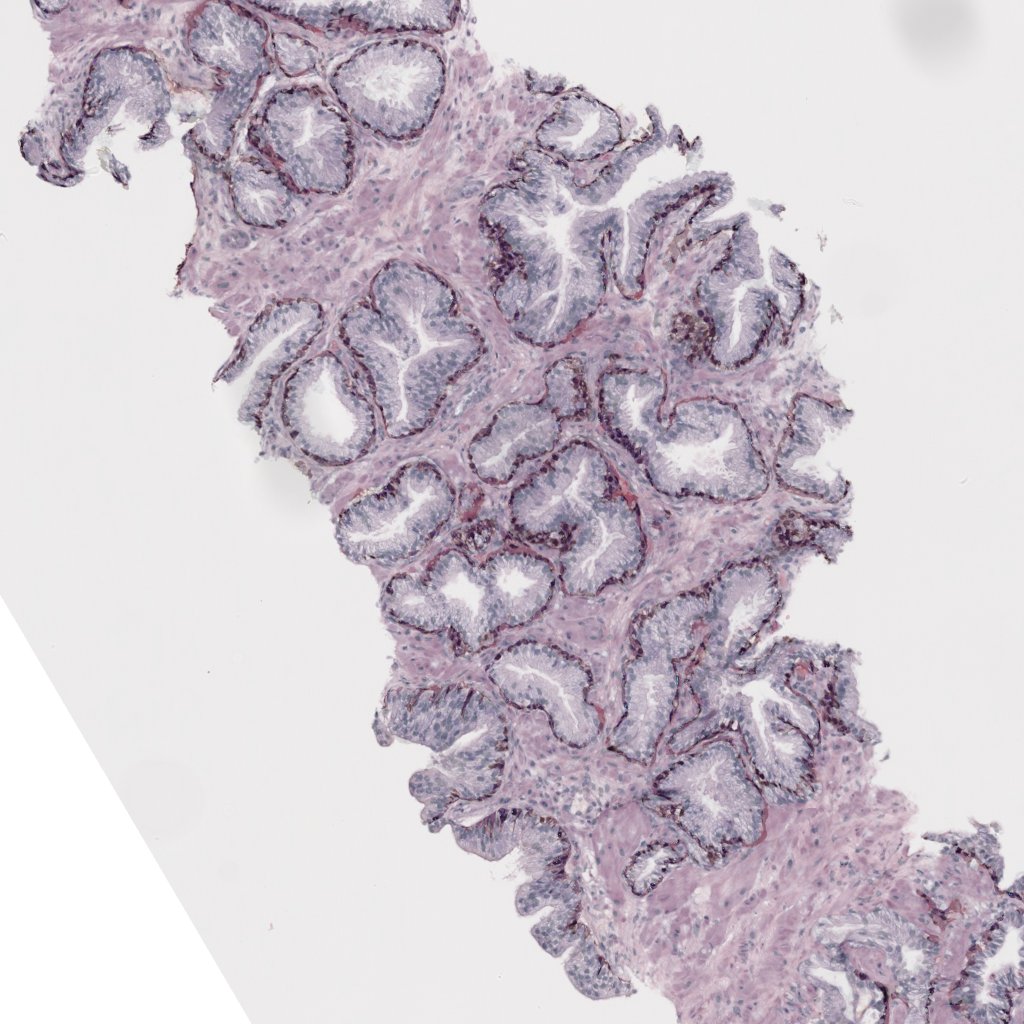}} \\[0.075\linewidth]
 & \tiny 1 & \tiny 2 & \tiny 3 & \tiny 4 & \tiny 5 \\
\end{tabular}
\caption{PIN-4 core images overlaid on H\&E core images at registration quality scores 1--5. A and B show independent examples at each score level. Scores of 4 or 5 were required for inclusion.}
\label{fig:qualitative_image_quality_examples}
\end{figure}
\endgroup
\subsection{Patch Extraction}
Registered H\&E and PIN-4 core images were loaded using TIAToolbox and paired $1024 \times 1024$ pixel patches were extracted from each registered core pair. Patches overlapping any Sedeen Viewer exclusion region were rejected, as were patches containing less than 5\% tissue by pixel count. Tissue was identified in HSV color space, where pixels satisfying $S \geq 25$ or $V \leq 200$ were classified as tissue; these thresholds were determined empirically. Rejection of a patch from either stain resulted in rejection of the corresponding patch in the paired core to preserve spatial correspondence. Before export, pure-white background pixels (all three RGB channels equal to 255) introduced during masking and registration were replaced with a patch-specific average background color, calculated independently per RGB channel from non-tissue background pixels. Tissue pixels were not modified. Processed patches were organized by stain type, tissue block, and core identifier.

\subsection{Patch Dataset Creation}
\begin{figure}[htbp]
    \centering
    \includegraphics[width=1\linewidth]{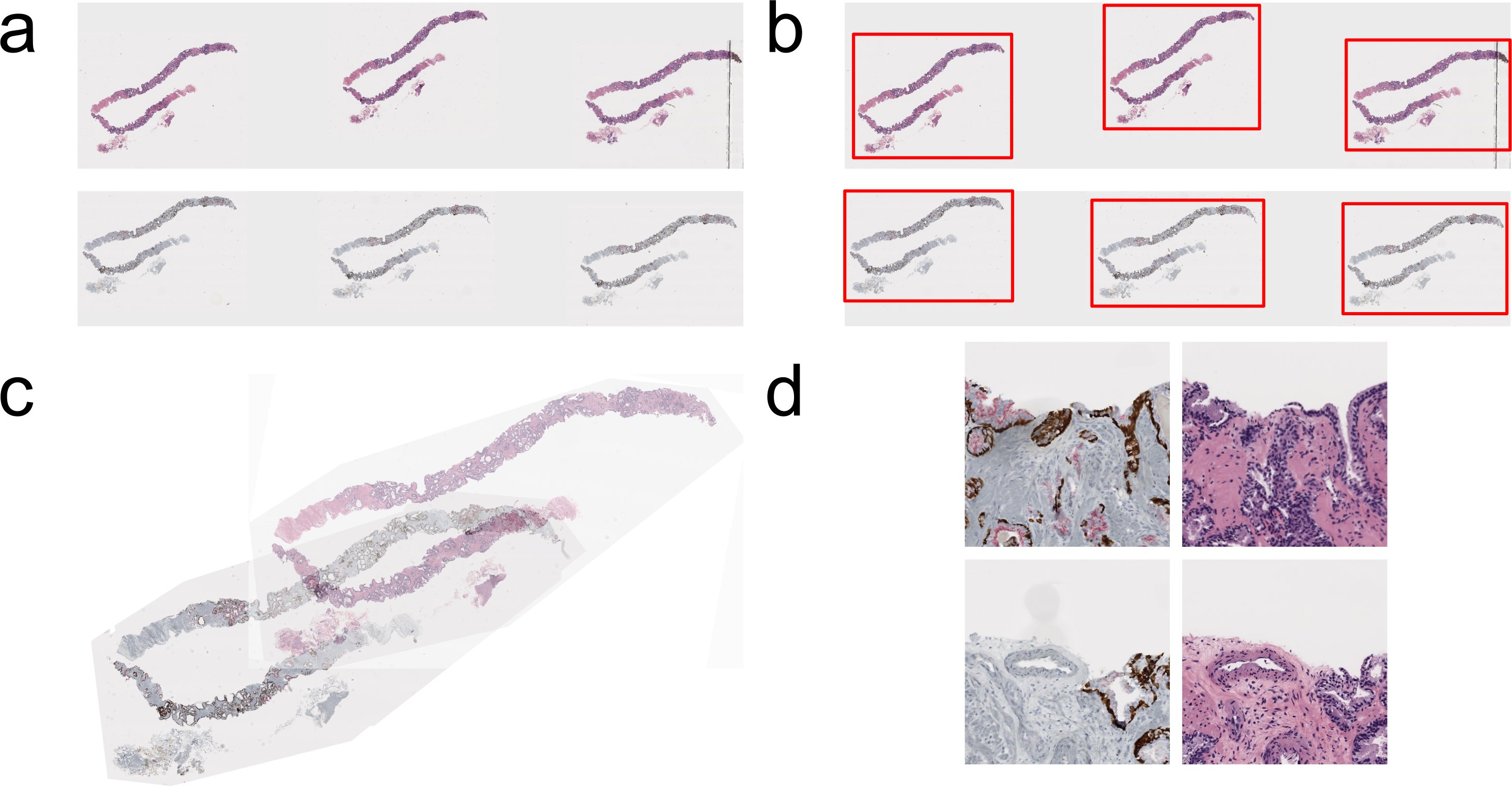}
    \caption{WSI preprocessing and patch extraction pipeline. (\textbf{a}) H\&E (top row) and PIN-4 (bottom row) core biopsy slides digitized at $20\times$ magnification. (\textbf{b}) H\&E (top row) and PIN-4 (bottom row) WSI pairs spatially registered for co-localization. (\textbf{c}) Registered core images aligned for paired patch extraction. (\textbf{d}) Representative paired $1024 \times 1024$ pixel patches extracted from regions with at least 5\% tissue coverage; PIN-4 (left column) and H\&E (right column).}
    \label{fig:preprocessing-pipeline}
\end{figure}

Only patches derived from core pairs with both image quality and registration quality scores of at least 4 were included in the final dataset. The resulting dataset contained 27,298 paired patches from 172 paired WSIs across 93 patients (Table~\ref{table:filtered_dataset_distribution}). The dataset was partitioned at the tissue-block level into training, validation, and held-out test sets using an approximate 80\%, 10\%, and 10\% split. All patches from the same tissue block were assigned to the same partition to prevent patch-level leakage. The final split contained 138 training blocks, 17 validation blocks, and 17 test blocks, yielding 22,576 training patches, 2,908 validation patches, and 1,814 test patches. Representation of adenocarcinoma-positive and adenocarcinoma-negative material was maintained across all three partitions. The preprocessing and patch extraction workflow is summarized in Fig.~\ref{fig:preprocessing-pipeline}, and described in detail in the following subsections.

\begin{table}[htbp]
\caption{Distribution of the paired dataset.}
\centering
\begin{minipage}{\linewidth}
\centering
\begin{NiceTabular}{l c c c}
 \toprule
  & Patients & WSI & Patches \\
 \midrule
 Count & 93 & 172 & 27298 \\
 \midrule
 \textbf{Biopsy} \\
 AC+\textsuperscript{*} & 54 (58\%) & 88 (51\%) & 15934 (58\%) \\
 AC--\textsuperscript{*} & 50 (54\%) & 84 (49\%) & 11364 (42\%) \\
 \midrule
 \textbf{Age} \\
 31-50 & 3 (3.2\%) & 5 (2.9\%) & 551 (2.0\%) \\
 51-70 & 64 (69\%) & 115 (67\%) & 17760 (65\%) \\
 71+ & 26 (28\%) & 52 (30\%) & 8987 (33\%) \\
 \midrule
 \textbf{Race} \\
 White & 62 (67\%) & 122 (71\%) & 20164 (74\%) \\
 Asian & 11 (12\%) & 19 (11\%) & 2667 (9.8\%) \\
 Black/AA\textsuperscript{\dag} & 1 (1.1\%) & 2 (1.2\%) & 206 (0.8\%) \\
 Other/Mixed & 10 (11\%) & 17 (9.9\%) & 2659 (9.7\%) \\
 Unknown & 9 (9.7\%) & 12 (7.0\%) & 1602 (5.9\%) \\
 \midrule
 \textbf{Ethnicity} \\
 Hispanic\textsuperscript{\ddag} & 17 (18\%) & 29 (17\%) & 4876 (18\%) \\
 Not Hispanic\textsuperscript{\ddag} & 75 (81\%) & 142 (83\%) & 22359 (82\%) \\
 Unknown & 1 (1.1\%) & 1 (0.6\%) & 63 (0.2\%) \\
 \bottomrule
\end{NiceTabular}
\par\smallskip
\justifying
\scriptsize
WSI and patch counts represent the number of registered H\&E/PIN-4 image pairs. Patients with both adenocarcinoma-positive and benign WSIs appear in both groups. \textsuperscript{*}AC+: adenocarcinoma-positive; AC$-$: adenocarcinoma-negative. \textsuperscript{\dag}Black or African American. \textsuperscript{\ddag}Hispanic, Latino, or Spanish origin.
\end{minipage}
\label{table:filtered_dataset_distribution}
\end{table}

\subsection{Pix2Pix Model}

\subsubsection{Architecture}

A cGAN based on the Pix2Pix architecture was adapted from a published and clinically benchmarked computational H\&E staining model \cite{rana_use_2020} and trained to learn the H\&E-to-PIN-4 image patch mapping. The model accepts a native $1024 \times 1024$ H\&E RGB patch acquired at $20\times$ magnification as input and generates a computationally stained PIN-4 RGB patch at the same spatial resolution. Initial experiments using transpose convolutions in the upsampling layers produced checkerboard artifacts. To mitigate these artifacts, stride-2 transpose convolution layers were replaced with nearest-neighbor upsampling by a factor of 2 followed by stride-1 convolution layers \cite{odena_deconvolution_2016}. The generator loss was also expanded to include a SSIM loss in addition to L1 and PCC losses \cite{wang_image_2004} to improve structural fidelity and color correspondence between generated and target PIN-4 patches. The loss functions were defined as:
{\small
\begin{align*}
  \mathcal{L}_{\text{cGAN}}(G, D)
    &= \mathbb{E}_{x,y}[\log D(x, y)] \\
    &\quad + \mathbb{E}_{x,y,z}[\log(1 - D(x, G(x,z)))], \\
  \mathcal{L}_1(G)
    &= \mathbb{E}_{x,y,z}[\|y - G(x,z)\|_1], \\
  \mathcal{L}_{\text{PCC}}(G)
    &= \mathbb{E}_{x,y,z}[1 - \text{PCC}(y,\, G(x,z))], \\
  \mathcal{L}_{\text{SSIM}}(G)
    &= \mathbb{E}_{x,y,z}[1 - \text{SSIM}(y,\, G(x,z))].
\end{align*}
\vspace{-2.3em}
\begin{align*}
  G^* &= \underset{G}{\arg\min}\;\underset{D}{\max}\;
  \mathcal{L}_{\text{cGAN}}(G, D)
  + \lambda\,\mathcal{L}_1(G) \\
  &\quad + \gamma\,\mathcal{L}_{\text{PCC}}(G)
  + \zeta\,\mathcal{L}_{\text{SSIM}}(G),
\end{align*}
where $x$ and $y$ are the input H\&E and target PIN-4 patches, respectively, and $z$ represents stochastic variation introduced through dropout. $\mathcal{L}_{\text{cGAN}}$ is the adversarial loss, and $\mathcal{L}_1$, $\mathcal{L}_{\text{PCC}}$, and $\mathcal{L}_{\text{SSIM}}$ are the L1, PCC, and SSIM losses between the generated and target PIN-4 images. The coefficients $\lambda$, $\gamma$, and $\zeta$ weight the L1, PCC, and SSIM loss terms, respectively.

\subsubsection{Hyperparameter Tuning}
Loss-function coefficients $\lambda$ (L1), $\gamma$ (PCC), and $\zeta$ (SSIM) were tuned using Ray Tune version 2.53.0 with Hyperopt version 0.2.7 and the Tree-structured Parzen Estimator algorithm \cite{liaw_tune_2018, bergstra_making_2013}, searching over ranges $[50, 150]$, $[1, 30]$, and $[1, 30]$, respectively. During tuning, models were trained for 20 epochs on a 25\% subset of the training data with a learning rate of $2 \times 10^{-4}$. The tuning objective minimized validation LPIPS \cite{zhang_unreasonable_2018} across 76 independent evaluations, yielding $\lambda = 61$, $\gamma = 20$, and $\zeta = 4.2$. With these coefficients fixed, nearest-neighbor upsampling followed by convolution outperformed bilinear upsampling across PCC, SSIM, and PSNR while reducing checkerboard artifacts. Training stability was evaluated over 30 epochs at learning rates of $5 \times 10^{-5}$, $10^{-4}$, and $2 \times 10^{-4}$; a rate of $10^{-4}$ provided the most stable trajectory.

\subsubsection{Training}
Models were trained using CUDA acceleration on an NVIDIA GeForce RTX 4080 GPU with 16\,GB of VRAM (NVIDIA, Santa Clara, CA). The AMSGrad variant of the Adam optimizer was used \cite{kingma_adam_2017, reddi_convergence_2019}. Random image flips and $90^\circ$ rotations were applied for data augmentation. The selected model was trained for 60 epochs using $\lambda = 61$, $\gamma = 20$, $\zeta = 4.2$, nearest-neighbor upsampling, and a learning rate of $10^{-4}$ for both generator and discriminator. Model checkpoints were saved after each epoch. The training workflow is summarized in Fig.~\ref{fig:model-training-diagram}.

\begin{figure}[ht]
    \centering
    \resizebox{\linewidth}{!}{
        \input{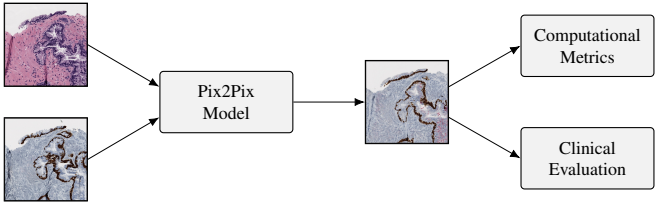}
    }
\caption{Pix2Pix model training and evaluation workflow. Input H\&E patches (top) and corresponding registered PIN-4 patches (bottom) from the training set were used to train the model. Validation patches guided iterative hyperparameter and architectural tuning via image reconstruction metrics. The best-performing model was evaluated on a held-out test set using PSNR, SSIM, LPIPS, and PCC, with additional qualitative pathologist review.}
\label{fig:model-training-diagram}
\end{figure}
\section{Results}

\subsection{Registration Quality Evaluation}

VALIS registration produced 974 paired H\&E/PIN-4 cores from 207 registered WSIs. Objective registration assessment showed strong tissue-mask overlap, with a median Dice index of 0.937 and a left-skewed distribution (Fig.~\ref{fig:registration_dice_index}). After manual review using the predefined image-quality and registration-quality rubrics, 722 of 974 registered core pairs (74.1\%) met inclusion criteria and were retained for patch extraction and model development. Representative examples of retained and excluded registrations are shown in Fig.~\ref{fig:qualitative_image_quality_examples}.

\begin{figure}[htbp]
    \centering
    \input{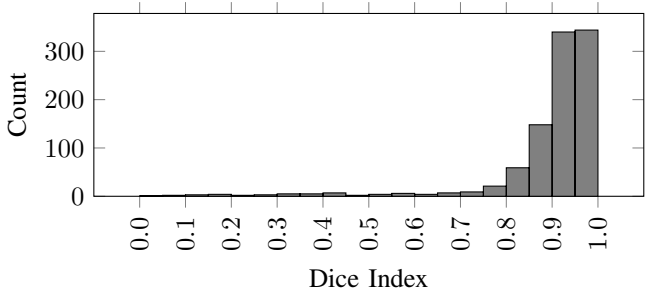}
    \caption{Dice index distribution for all registered core pairs. Tissue masks were generated by converting RGB pixels to HSV color space and classifying pixels satisfying $S \geq 10$ or $V \leq 225$ as tissue. The distribution is left-skewed with a median Dice index of 0.937, indicating strong spatial alignment.}
    \label{fig:registration_dice_index}
\end{figure}

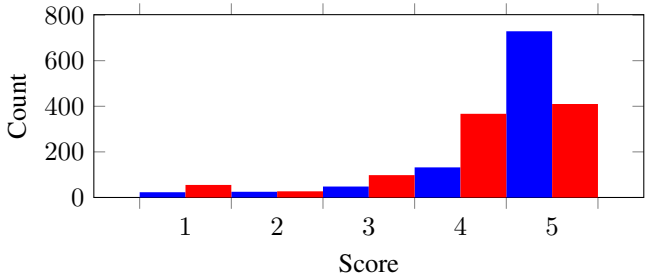
\begin{figure}[htbp]
    \centering
    \begin{tikzpicture}
  \begin{axis}[
    ybar interval,
    xlabel={Score},
    ylabel={Count},
    ymin=0,
    xtick={1, 2, 3, 4, 5, 6},
    xticklabel pos=left,
    grid=none,
    width=\linewidth,
    height=4cm
  ]
    \addplot[
      fill=blue,
      draw=none,
      hist={bins=5, data min=1, data max=6}
    ] table[y index=0] {
      data
5
5
5
1
5
5
5
5
5
4
4
5
4
4
5
5
5
5
3
5
5
2
4
5
5
5
4
5
5
2
5
3
4
3
4
4
4
5
5
5
5
5
5
5
5
5
4
5
5
5
4
5
5
4
2
5
3
4
5
5
2
5
5
5
5
5
5
5
5
5
5
5
5
5
5
5
5
2
3
3
2
4
4
3
5
5
5
5
5
5
5
5
5
5
5
5
5
5
4
5
5
5
5
5
5
5
4
5
5
5
5
5
4
5
5
1
2
4
1
2
5
5
5
4
5
4
5
5
5
5
4
5
3
5
4
4
5
5
5
5
1
3
1
3
1
3
3
1
2
3
2
2
2
3
5
5
5
5
5
5
5
5
5
5
5
5
5
2
5
5
5
5
5
5
5
5
5
5
5
5
5
5
5
5
5
5
4
5
5
5
5
5
5
5
5
5
5
4
5
3
3
4
3
3
3
3
3
3
5
3
3
3
3
3
5
5
5
5
5
5
5
5
5
5
5
5
5
4
5
4
5
5
5
4
5
4
5
4
4
4
5
5
5
5
5
5
5
5
4
5
5
5
4
5
5
4
3
3
5
4
5
5
5
4
5
5
5
5
5
2
1
1
2
5
5
5
4
4
5
5
5
5
5
5
5
3
5
5
5
5
5
5
5
5
5
5
5
5
5
5
5
5
5
5
5
5
5
3
5
5
4
5
5
5
4
4
5
5
5
5
5
5
5
5
3
5
5
5
5
5
1
3
5
3
4
5
3
5
4
5
5
5
2
3
3
5
1
1
4
4
5
5
4
5
5
5
5
5
5
5
5
5
5
5
5
5
5
5
5
3
5
5
5
5
5
5
5
4
4
4
5
2
4
5
5
5
4
5
5
4
5
5
5
5
5
5
4
5
5
5
5
5
4
5
5
1
5
1
5
5
5
5
4
4
3
4
4
4
5
5
4
4
4
5
4
5
5
5
4
5
5
5
5
5
4
5
5
5
4
5
5
5
1
5
5
5
5
5
5
5
5
5
5
5
5
5
5
5
5
5
4
5
5
1
1
1
1
1
5
5
5
3
5
4
5
5
5
5
5
4
5
5
5
4
5
4
5
3
5
5
5
5
5
4
5
5
5
5
5
5
5
5
5
5
5
5
5
5
5
5
5
5
5
5
5
5
5
5
5
5
5
5
5
5
5
5
5
5
5
5
4
4
4
4
4
5
5
5
5
4
5
5
5
5
4
4
4
5
5
5
5
5
5
5
5
5
5
5
5
4
5
4
5
5
5
5
5
5
5
5
5
5
5
5
4
2
4
5
4
4
4
5
5
5
4
4
4
5
4
5
5
5
2
4
4
5
4
5
5
5
4
5
5
5
4
5
5
4
5
5
5
4
4
5
5
5
5
5
5

2
2
4
5
5
5
5
5
5
2
2
5
5
5
4
5
5
5
5
5
5
4
5
5
5
5
5
5
5
5
4
5
5
5
5
5
1
3
1
4
4
3
4
1
5
5
3
3
4
5
5
5
5
4
4
5
5
5
5
5
5
5
5
5
3
5
5
5
5
5
5
5
5
5
5
5
5
5
5
4
5
4
5
5
4
5
5
4
5
5
5
5
5
5
5
5
5
5
5
5
5
5
5
5
5
5
5
5
5
4
5
5
5
5
5
5
3
5
5
5
5
5
5
5
5
5
5
5
5
5
5
5
5
5
5
5
5
5
5
5
5
5
5
5
4
5
4
5
5
5
5
5
5
5
5
5
5
5
5
4
5
5
5
5
5
5
4
5
5
5
5
5
4
5
4
5
4
5
5
5
5
5
3
4
4
2
5
5
4
5
5
5
5
5
5
5
5
5
2
5
5
5
5
5
5
5
5
5
5
5
5
5
5
4
5
5
5
5
5
4
5
5
5
5
5
5
5
5
5
5
5
5
5
5
5
5
5
5
5
5
5
5
5
5
5
5
5
5
5
5
5
4
5
5
5
5
5
5
5
5
5
5
5
5
5
5
5
5
5
5
5
5
5
5
5
5
5
5
5
5
5
5
5
5
5
5
5
5
5
5
5
5
5
5
5
5
5
5
5
5
5
5
5
5
5
5
5
5
5
5
5
5
5
5
5
5
5
5
5
5
5
5
5
5
5
5
5
5
5
5
5
5
5
5
5
5
5
5
    };
    \addplot[
      fill=red,
      draw=none,
      hist={bins=5, data min=1, data max=6}
    ] table[y index=0] {
      data
3
3
4
1
5
4
5
5
3
3
1
3
2
2
3
3
1
4
2
4
3
4
3
4
4
4
4
1
4
1
3
1
4
1
4
4
4
4
3
4
4
5
4
3
4
4
2
3
4
5
4
4
4
2
4
5
3
3
2
4
1
4
4
3
4
4
4
1
4
4
5
3
3
1
3
4
5
2
1
1
3
4
4
1
4
4
4
4
4
5
3
5
3
5
4
5
4
5
4
4
4
5
4
4
3
4
3
3
4
3
5
3
3
5
5
3
4
4
3
4
4
5
4
5
4
4
3
4
4
4
4
4
4
3
4
4
5
5
5
5
1
5
1
5
3
5
3
1
4
4
4
4
4
4
1
5
3
1
4
4
3
4
4
4
3
2
1
4
5
5
3
3
4
3
4
3
2
1
4
4
3
2
4
5
3
4
4
4
4
4
4
3
5
4
3
4
5
4
5
5
4
4
4
4
4
4
4
4
3

4
4
4
5
5
5
4
2
5
1
5
1
4
5
4
4
3
4
4
5
4
4
4
4
3
5
2
5
5
3
4
5
5
5
5
4
5
5
5
4
4
4
3
4
4
5
5
5
5
5
5
3
5
5
4
5
5
4
5
5
1
4
1
1
4
4
5
4
5
4
5
5
3
4
4
4
2
4
4
4
3
4
4
1
4
3
1
5
5
4
3
1
3
3
4
4
3
3
5
5
4
4
4
4
4
3
5
4
5
3
3
4
4
4
4
3
5
5
4
5
4
3
5
5
4
4
4
4
4
4
5
1
4
2
5
4
4
2
4
5
4
5
5
4
4
4
3
4
3
4
5
4
4
3
4
4
5
4
5
5
5
4
4
5
5
5
5
5
2
5
2
5
3
3
5
1
5
5
5
5
3
4
4
5
5
5
5
5
5
4
5
5
5
5
5
5
2
5
3
5
5
5
5
4
5
5
5
5
5
5
5
5
5
5
5
5
5
5
5
5
5
5
5
5
5
5
1
5
1
3
4
3
4
3
3
5
5
4
5
5
5
4
4
4
4
5
4
4
4
4
4
4
4
5
3
1
3
3
3
5
3
5
3
5
3
5
5
4
4
4
4
4
4
4
2
5
2
5
2
5
5
5
5
5
5
4
3
3
4
4
5
4
5
4
5
5
5
5
4
5
4
4
5
5
4
5
5
5
5
4
4
4
4
4
4
4
4
4
5
5
5
5
4
4
4
4
5
5
4
5
5
5
4
5
5
5
4
4
5
5
5
4
5
5
5
5
5
5
5
5
4
4
5
4
4
4
1
4
4
4
4
4
5
3
1
5
5
4
4
5
5
5
5
1
1
4
5
5
4
4
4
5
1
4
5
5
4
4
5
4
5
4
4
5
4
4
4
4
4
5
4
4
5
5
5
4
5
4
4
4

5
5
5
4
4
4
5
4
4
5
5
4
5
4
5
2
4
4
1
4
4
4
4
4
5
5
4
5
5
4
4
5
4
4
4
4
5
1
5
4
5
4
5
4
5
4
5
2
2
5
4
5
5
5
5
4
5
4
5
5
5
4
5
5
4
4
5
5
5
5
4
4
1
5
5
4
4
5
5
4
5
4
5
4
4
5
5
5
5
5
3
5
5
5
4
4
5
4
4
5
4
4
4
3
4
4
4
1
3
5
5
4
4
5
5
5
5
5
4
5
5
5
5
4
5
5
5
5
5
5
5
4
5
5
5
5
5
5
5
4
5
5
5
5
4
5
4
1
1
1
5
5
4
5
5
5
5
5
5
5
5
5
5
5
5
4
4
5
5
5
4
5
4
5
4
5
4
4
4
5
5
4
5
3
5
4
4
4
2
4
4
4
4
3
4
1
4
2
1
4
5
4
5
5
5
5
5
5
5
4
4
5
4
5
4
5
4
4
5
3
5
3
4
5
4
5
4
4
5
5
5
5
5
5
4
5
5
5
5
5
5
5
5
5
5
5
5
5
5
5
4
5
5
5
5
3
5
5
5
4
5
5
4
5
5
5
5
5
5
4
5
5
5
5
5
5
4
5
4
5
5
5
4
5
5
5
5
5
5
5
5
3
5
3
1
1
4
1
1
4
1
5
2
5
4
5
5
5
5
5
5
5
5
4
4
5
4
4
5
5
5
5
5
5
5
5
4
4
4
4
3
3
1
1
3
1
4
4
    };
  \end{axis}
\end{tikzpicture}
    \caption{Distribution of qualitative evaluation scores for registered cores. Blue and red bars represent image quality and registration scores, respectively. Assessment was performed using a five-point rubric to evaluate image quality and registration accuracy.}
    \label{fig:subjective_evaluation_scores}
\end{figure}

\subsection{H\&E to PIN-4 Computational Staining Model Evaluation}
 
\subsubsection{Model Checkpoint Selection}

Model checkpoints were evaluated at each epoch by reconstructing a 25\% subset of validation cores from generated patches to assess PIN-4 synthesis quality. The epoch 60 checkpoint was selected for final computational and clinical evaluation based on the best qualitative balance of tissue clarity, racemase localization, basal-cell-associated staining, and limited false-positive staining in benign-appearing regions.

\subsubsection{Quantitative Benchmarking of the Baseline Model}
The selected Pix2Pix model checkpoint at epoch 60 was evaluated across training, validation, and held-out test partitions (Table~\ref{table:test-metrics}). On the training set, the model achieved a mean PCC of 0.667, SSIM of 0.662, PSNR of 21.72\,dB, and LPIPS of 0.411. On the validation set, the model achieved a mean PCC of 0.651, SSIM of 0.657, PSNR of 21.88\,dB, and LPIPS of 0.412. On the held-out test set, the model achieved a mean PCC of 0.684, SSIM of 0.667, PSNR of 21.88\,dB, and LPIPS of 0.417. Performance was similar across the three partitions, suggesting that the model generalized to held-out tissue blocks without substantial degradation in image-similarity performance. Mean per-channel RGB absolute differences between generated and ground-truth patches were small across WSIs, with median $|$Diff.$|$ values of 3.2, 3.6, and 3.4 for the red, green, and blue channels, respectively (Table~\ref{table:rgb-metrics}). These results indicate close overall color correspondence between generated and native PIN-4 images, while recognizing that RGB channel agreement does not by itself establish marker-level diagnostic accuracy. Per-WSI image-generation metrics for the validation and test partitions are provided in the supplementary material (Table~\ref{table:test-metrics}).

\subsection{Pathologist Review of Generated PIN-4 Images}
Generated PIN-4 images were reviewed alongside the corresponding source H\&E and native PIN-4 targets by a board-certified pathologist. Review focused on AMACR/racemase-like signal, basal-cell-associated staining, glandular architecture, and discordant regions between synthetic and native PIN-4 images. Representative examples are shown in Fig.~\ref{fig:gen-patch-examples}. In strong-performing examples, computationally stained PIN-4 images showed visual correspondence with the approximate distribution of AMACR/racemase-like signal in suspicious or malignant-appearing glands while preserving glandular architecture visible on H\&E. Benign-appearing glands frequently demonstrated basal-cell-associated staining patterns. In intermediate examples, generated images preserved overall tissue structure but showed weaker or less complete marker localization compared with native PIN-4. Areas of discordance included incomplete AMACR/racemase-like signal in some malignant-appearing regions, variability in basal-cell-associated staining, and occasional marker-like signal outside expected epithelial compartments. Representative outlier cases included intraductal carcinoma, where the model generated recognizable staining context sufficient for pathologist interpretation despite morphologic complexity, and high-grade carcinoma, where extensive tissue disruption produced morphology underrepresented in the training set, resulting in incomplete marker synthesis.
\begin{figure}[htbp]
    \centering
    \includegraphics[width=0.8\linewidth]{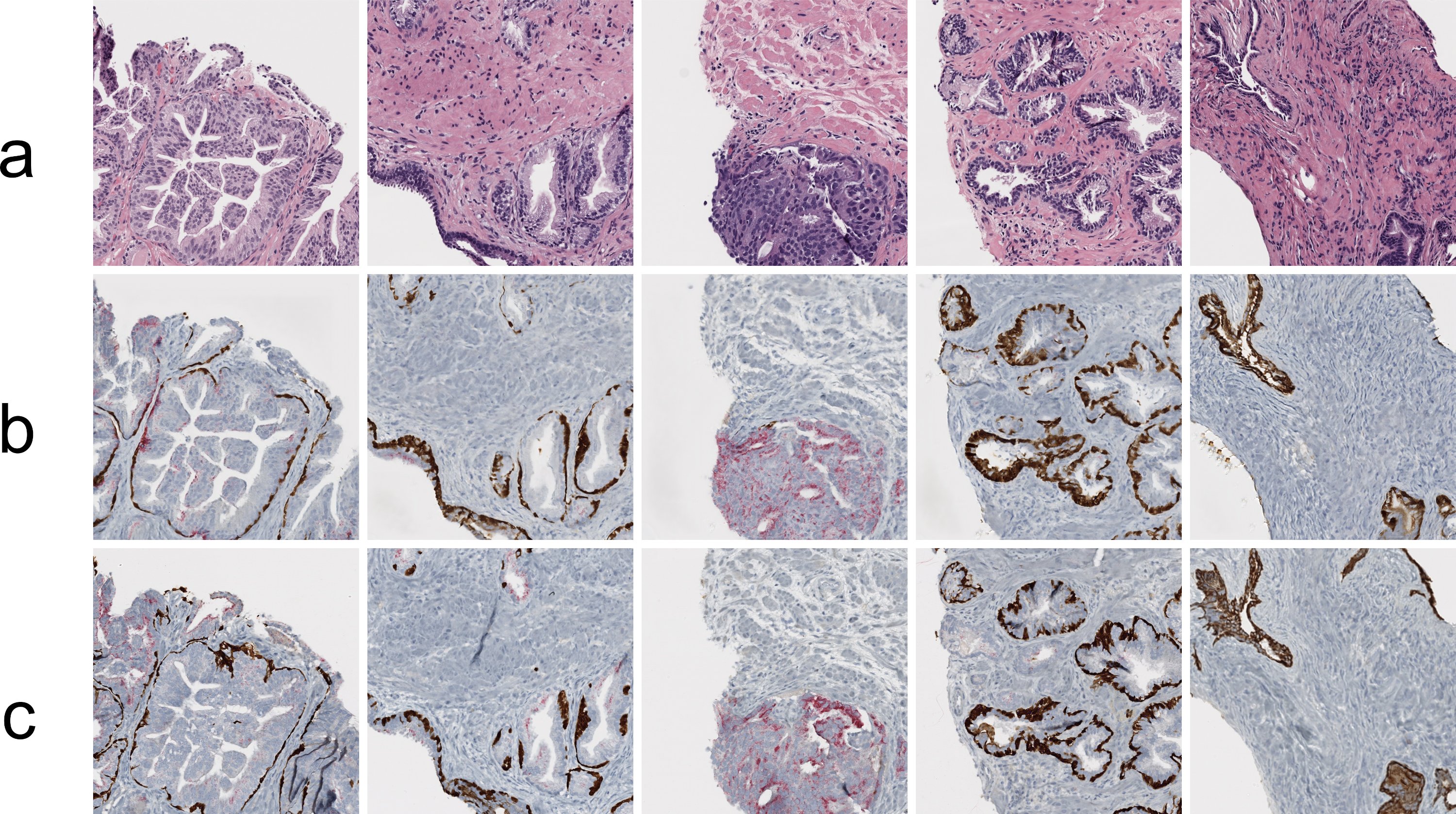}
    \caption{Representative examples of generated PIN-4 patches displayed as $1024 \times 1024$ pixel patches at $20\times$ magnification. Each column shows one example: (\textbf{a}) source H\&E patch, (\textbf{b}) generated PIN-4 patch, and (\textbf{c}) corresponding ground truth PIN-4 patch. The rightmost two columns show examples from WSIs containing adenocarcinoma.}
    \label{fig:gen-patch-examples}
\end{figure}
\section{Discussion}

The dataset composition supports the feasibility of supervised PIN-4 immunophenotypic pattern synthesis from routinely acquired brightfield H\&E prostate biopsy images. The cohort was derived from routine clinical material rather than a narrowly selected experimental dataset, reflecting the tissue variability, staining variation, and morphologic heterogeneity encountered in clinical prostate biopsy workflows. The dataset included patients across age, race, and ethnicity groups (Table~\ref{table:filtered_dataset_distribution}). The inclusion of both adenocarcinoma-positive and adenocarcinoma-negative material required the model to learn differential morphology-to-marker relationships such as generating AMACR/racemase-like signal in suspicious glandular regions while preserving basal-cell-associated staining in benign-appearing glands and avoiding nonspecific carcinoma-like signal.

The registration pipeline achieved sufficient spatial correspondence for supervised H\&E-to-PIN-4 learning despite tissue deformation, sectioning-depth differences, and scanning variation inherent to serial-section histopathology. Beyond automated alignment, the downstream quality-control process contributed meaningfully to core-level review. For example, manual exclusion of misregistered or artifact-containing regions (Fig.~\ref{fig:qualitative_image_quality_examples}) reduced noisy patch pairs that would otherwise degrade training signal. This was particularly important for prostate biopsy material, where individual cores frequently contain focal discontinuities, folds, missing tissue, or local deformation. The combination of automated registration and manual quality control produced a dataset in which H\&E morphology and PIN-4 signal are reliably co-localized, a prerequisite for learning meaningful morphology-to-marker mappings.

Pathologist review provided clinically meaningful context for interpreting the quantitative results (Table~\ref{table:test-metrics}, Fig.~\ref{fig:gen-patch-examples}). Prostate cancer diagnosis relies on glandular architecture assessed on H\&E, where tissue morphology is the diagnostic gold standard \cite{delahunt_gleason_2012, humphrey_histopathology_2017}. PIN-4 IHC is ordered to interrogate that architecture, confirming or excluding malignancy based on AMACR/racemase expression and basal cell presence in the same glands raising diagnostic concern \cite{sabata_automated_2010}. Conventional adjacent-section IHC cannot guarantee that the glands assessed on PIN-4 are the same glands seen on H\&E, due to tissue loss, deformation, and sectioning depth differences. The key question in this study was therefore not whether the synthetic image perfectly matched the native PIN-4 slide, but whether the generated staining patterns preserved plausible spatial relationships with the corresponding H\&E morphology that would support pathologist interpretation of the same tissue. The qualitative findings demonstrated that the model outputs captured several of these relationships. Correspondence between generated AMACR/racemase-like signal and suspicious or malignant-appearing glands, combined with preservation of basal-cell-associated staining in benign-appearing glands (Fig.~\ref{fig:gen-patch-examples}), indicated that the model learned morphologically grounded associations rather than color statistics. This is the minimum requirement for the output to serve as a useful co-localized visual adjunct alongside the source H\&E morphology. Model errors including variability in basal-cell-associated staining, incomplete racemase-like signal in some malignant regions, and occasional marker-like signal outside expected epithelial compartments were observed predominantly in morphologically complex regions such as intraductal carcinoma and high-grade carcinoma, where training examples were limited. These findings position the generated images as a potential adjunctive co-localization tool to review predicted PIN-4 marker patterns in spatial context with the source H\&E architecture, rather than as a standalone diagnostic stain.

The quantitative results show that the held-out test set achieved a PSNR of 21.88\,dB, SSIM of 0.667, PCC of 0.684, and LPIPS of 0.417 (Table~\ref{table:test-metrics}). PSNR and SSIM reflect pixel-level and structural fidelity to the registered native PIN-4 target, which is an imperfect reference because serial-section registration introduces residual tissue differences that limit the maximum achievable similarity regardless of model quality. PCC was higher than SSIM, reflecting stronger preservation of the spatial distribution of marker signal than fine staining texture, consistent with a model learning morphology-to-marker associations rather than exact pixel reconstruction. Despite these constraints, the SSIM of 0.667 exceeded values reported for H\&E-to-HER2 translation in breast tissue \cite{peng_advancing_2024} and H\&E-to-GPC3 translation in liver tissue \cite{li_virtual_2024}, which reported SSIM values of approximately 0.34 and 0.458 respectively. These comparisons should be interpreted cautiously given differences in tissue type, stain target, and dataset, but they suggest that PIN-4 staining patterns may be more learnable from H\&E morphology than HER2 or GPC3 markers in other tissue types. The prior prostate PIN-4 model relied on hyperspectral autofluorescence microscopy with approximately 20 spectral channels versus the 3 RGB channels used here \cite{wong_clinical-grade_2024}, providing substantially richer input information; that the present model achieved comparable qualitative PIN-4 synthesis using only standard brightfield H\&E input is therefore a meaningful result. The stability of SSIM and PSNR across training, validation, and test partitions confirmed generalization to held-out tissue blocks without substantial performance degradation. The narrow interquartile ranges for per-channel RGB absolute differences (Table~\ref{table:rgb-metrics}) support generally consistent color correspondence across WSIs. RGB channel differences reflect overall color fidelity rather than marker-level accuracy, providing complementary information to structural and perceptual image-similarity metrics. The moderate association between RGB differences and image-generation metrics, including SSIM ($r = -0.46$) and LPIPS ($r = 0.39$), suggests that color-channel agreement and perceptual similarity capture related but distinct aspects of generated image quality.

Although this study included a relatively large, paired H\&E/PIN-4 patch dataset from multiple patients, it was retrospective and single-institutional. Future work should expand the dataset with additional cases from multiple institutions to assess generalizability across staining protocols, scanners, tissue processing workflows, and patient populations. A larger pathologist reader study with predefined diagnostic endpoints will also be necessary to evaluate the potential role of computational PIN-4 in specific clinical use cases. WSI-level metric aggregation and training datasets enriched for rare or diagnostically challenging prostate lesions, including intraductal carcinoma and atypical small acinar proliferation, are additional growth areas. The study contributes a reproducible workflow for constructing paired, registered H\&E/IHC datasets from routine clinical WSIs, which may support future computational staining studies in prostate pathology and other tissue types. Upon prospective clinical validation, computational PIN-4 could allow pathologists to inspect predicted AMACR/racemase-like and basal-cell-associated staining patterns alongside the H\&E architecture, particularly in regions where serial-section differences or tissue loss complicate comparison with native IHC. Future generative model development could benefit from tissue-compartment-aware architectures, loss functions, and evaluation strategies that emphasize glandular epithelium over background and stromal areas, enabling more targeted assessment of computational H\&E-to-PIN-4 staining in diagnostically relevant clinical contexts.
\section*{Acknowledgment \& Code}
\vspace{-0.9em}
{\small Dr. Alarice Lowe performed pathology interpretations of model outputs and ground truth images. Code available at: \url{https://github.com/Dr-Pratik-Shah-UCI/genAI\_V\_IHC}}

\bibliographystyle{IEEEtran}
\bibliography{paper}

\IEEEtriggeratref{26}
\section*{Supplementary Material}
\begingroup
\setlength{\tabcolsep}{3pt}
\begin{table}[htbp]
\caption{Image generation metrics across 34 validation and test WSIs}
\centering
\scriptsize
\begin{NiceTabular}{r c c c c c}
\toprule
WSI & Patches & LPIPS $\downarrow$ & SSIM $\uparrow$ & PCC $\uparrow$ & PSNR $\uparrow$ \\
\midrule
1 & 63 & 0.573 & 0.315 & 0.267 & 14.953 \\
2 & 173 & 0.437 & 0.638 & 0.694 & 21.201 \\
3 & 12 & 0.392 & 0.798 & 0.675 & 23.875 \\
4 & 146 & 0.375 & 0.690 & 0.644 & 20.447 \\
5 & 172 & 0.381 & 0.730 & 0.718 & 24.063 \\
6 & 150 & 0.410 & 0.634 & 0.697 & 21.443 \\
7 & 50 & 0.418 & 0.676 & 0.728 & 21.810 \\
8 & 104 & 0.364 & 0.740 & 0.711 & 23.589 \\
9 & 140 & 0.418 & 0.664 & 0.743 & 22.606 \\
10 & 47 & 0.420 & 0.629 & 0.562 & 21.581 \\
11 & 23 & 0.357 & 0.753 & 0.548 & 21.798 \\
12 & 116 & 0.412 & 0.710 & 0.710 & 21.873 \\
13 & 116 & 0.411 & 0.705 & 0.764 & 24.513 \\
14 & 84 & 0.420 & 0.690 & 0.747 & 22.502 \\
15 & 208 & 0.454 & 0.677 & 0.702 & 22.384 \\
16 & 129 & 0.409 & 0.644 & 0.680 & 19.942 \\
17 & 81 & 0.426 & 0.642 & 0.660 & 21.373 \\
18 & 242 & 0.549 & 0.459 & 0.237 & 16.434 \\
19 & 109 & 0.386 & 0.661 & 0.697 & 22.506 \\
20 & 92 & 0.402 & 0.683 & 0.716 & 23.132 \\
21 & 77 & 0.305 & 0.790 & 0.733 & 22.533 \\
22 & 164 & 0.401 & 0.666 & 0.689 & 20.577 \\
23 & 224 & 0.420 & 0.692 & 0.699 & 24.511 \\
24 & 205 & 0.390 & 0.663 & 0.671 & 22.174 \\
25 & 381 & 0.432 & 0.581 & 0.577 & 22.217 \\
26 & 188 & 0.370 & 0.673 & 0.716 & 20.831 \\
27 & 227 & 0.402 & 0.634 & 0.663 & 20.783 \\
28 & 75 & 0.439 & 0.682 & 0.716 & 21.923 \\
29 & 369 & 0.382 & 0.711 & 0.738 & 23.143 \\
30 & 85 & 0.418 & 0.654 & 0.691 & 19.740 \\
31 & 136 & 0.398 & 0.755 & 0.782 & 25.692 \\
32 & 104 & 0.397 & 0.723 & 0.691 & 22.638 \\
33 & 147 & 0.393 & 0.718 & 0.700 & 22.485 \\
34 & 83 & 0.427 & 0.702 & 0.735 & 21.720 \\
\bottomrule
\end{NiceTabular}
\par\smallskip
\justifying
\scriptsize
LPIPS ranges from 0 to 1. SSIM and PCC range from 0 to 1. PSNR is in dB.
\label{table:test-metrics}
\end{table}
\endgroup
\begin{table*}[t]
\caption{Mean per-patch RGB channel values for generated and ground truth images by origin WSI}
\centering
\scriptsize
\setlength{\tabcolsep}{2.5pt}
\begin{minipage}[t]{0.32\textwidth}
\centering
\textbf{Red Channel}\\[2pt]
\begin{NiceTabular}{r c c c c}
\toprule
WSI & Gen. & GT & Diff. & $|$Diff.$|$ \\
\midrule
1 & 191.9$\pm$16.7 & 188.4$\pm$19.5 & 3.5$\pm$7.1 & 6.0$\pm$5.1 \\
2 & 199.4$\pm$24.9 & 198.5$\pm$27.6 & 0.9$\pm$5.2 & 3.9$\pm$3.6 \\
3 & 216.0$\pm$21.1 & 221.1$\pm$19.2 & -5.0$\pm$2.5 & 5.0$\pm$2.5 \\
4 & 204.9$\pm$20.7 & 209.6$\pm$20.5 & -4.7$\pm$3.5 & 5.0$\pm$3.1 \\
5 & 213.1$\pm$15.7 & 215.4$\pm$16.3 & -2.3$\pm$2.3 & 2.9$\pm$1.5 \\
6 & 200.8$\pm$22.7 & 199.3$\pm$25.8 & 1.4$\pm$5.1 & 3.9$\pm$3.6 \\
7 & 206.2$\pm$21.4 & 205.5$\pm$24.2 & 0.7$\pm$4.1 & 3.0$\pm$2.8 \\
8 & 213.2$\pm$17.5 & 214.3$\pm$19.7 & -1.2$\pm$3.4 & 3.0$\pm$1.9 \\
9 & 203.6$\pm$20.8 & 204.0$\pm$22.4 & -0.3$\pm$3.5 & 2.8$\pm$2.1 \\
10 & 205.4$\pm$21.8 & 208.1$\pm$22.6 & -2.6$\pm$2.8 & 3.4$\pm$1.9 \\
11 & 216.6$\pm$13.9 & 230.9$\pm$4.9 & -14.4$\pm$9.1 & 14.4$\pm$9.1 \\
12 & 210.7$\pm$18.4 & 210.2$\pm$21.9 & 0.5$\pm$4.7 & 2.9$\pm$3.7 \\
13 & 211.9$\pm$14.4 & 211.9$\pm$17.1 & 0.0$\pm$3.4 & 2.8$\pm$1.9 \\
14 & 205.9$\pm$19.0 & 210.3$\pm$18.2 & -4.3$\pm$2.9 & 4.5$\pm$2.6 \\
15 & 210.0$\pm$18.3 & 211.1$\pm$20.0 & -1.1$\pm$3.1 & 2.7$\pm$1.9 \\
16 & 202.0$\pm$22.4 & 207.2$\pm$21.9 & -5.2$\pm$3.7 & 5.4$\pm$3.4 \\
17 & 204.5$\pm$21.8 & 202.8$\pm$25.5 & 1.7$\pm$4.3 & 3.5$\pm$3.0 \\
18 & 197.2$\pm$15.5 & 194.0$\pm$16.9 & 3.2$\pm$6.3 & 5.3$\pm$4.8 \\
19 & 207.1$\pm$20.5 & 210.2$\pm$21.2 & -3.0$\pm$2.5 & 3.6$\pm$1.6 \\
20 & 210.2$\pm$16.6 & 210.8$\pm$18.2 & -0.7$\pm$3.9 & 3.1$\pm$2.4 \\
21 & 217.5$\pm$14.0 & 220.1$\pm$14.8 & -2.6$\pm$2.4 & 3.0$\pm$1.9 \\
22 & 203.6$\pm$21.2 & 202.2$\pm$24.5 & 1.4$\pm$4.3 & 3.4$\pm$3.0 \\
23 & 214.2$\pm$14.2 & 213.4$\pm$16.4 & 0.8$\pm$3.0 & 2.4$\pm$1.8 \\
24 & 212.8$\pm$15.2 & 211.7$\pm$18.2 & 1.1$\pm$4.0 & 3.3$\pm$2.5 \\
25 & 209.0$\pm$16.5 & 208.7$\pm$18.1 & 0.3$\pm$4.0 & 3.2$\pm$2.3 \\
26 & 205.4$\pm$19.3 & 206.5$\pm$21.6 & -1.1$\pm$3.4 & 2.9$\pm$2.1 \\
27 & 205.7$\pm$18.1 & 206.4$\pm$20.0 & -0.7$\pm$3.6 & 3.1$\pm$2.0 \\
28 & 209.1$\pm$17.4 & 203.5$\pm$23.7 & 5.6$\pm$7.2 & 6.4$\pm$6.5 \\
29 & 212.8$\pm$16.1 & 213.2$\pm$18.0 & -0.3$\pm$3.1 & 2.5$\pm$1.8 \\
30 & 202.8$\pm$22.7 & 204.7$\pm$23.8 & -1.9$\pm$2.3 & 2.7$\pm$1.4 \\
31 & 214.9$\pm$14.9 & 215.1$\pm$17.0 & -0.2$\pm$2.8 & 2.3$\pm$1.6 \\
32 & 209.9$\pm$16.4 & 217.0$\pm$14.1 & -7.1$\pm$3.7 & 7.1$\pm$3.6 \\
33 & 207.3$\pm$21.9 & 211.6$\pm$21.2 & -4.3$\pm$2.7 & 4.4$\pm$2.6 \\
34 & 208.9$\pm$19.8 & 207.8$\pm$22.8 & 1.1$\pm$4.3 & 3.1$\pm$3.1 \\
\midrule
Mean & 207.8$\pm$5.8 & 209.0$\pm$7.9 & -1.2$\pm$3.6 & 4.0$\pm$2.2 \\
Median & 208.1 & 209.9 & -0.5 & 3.2 \\
Q1     & 204.6 & 204.9 & -2.6 & 2.9 \\
Q3     & 212.6 & 212.9 &  0.9 & 4.5 \\
\bottomrule
\end{NiceTabular}
\end{minipage}
\hfill
\begin{minipage}[t]{0.32\textwidth}
\centering
\textbf{Green Channel}\\[2pt]
\begin{NiceTabular}{r c c c c}
\toprule
WSI & Gen. & GT & Diff. & $|$Diff.$|$ \\
\midrule
1 & 186.6$\pm$18.5 & 172.6$\pm$28.3 & 14.0$\pm$16.7 & 15.8$\pm$15.0 \\
2 & 200.2$\pm$23.3 & 198.8$\pm$26.3 & 1.4$\pm$5.4 & 4.1$\pm$3.8 \\
3 & 214.3$\pm$20.7 & 219.3$\pm$19.1 & -5.0$\pm$3.0 & 5.0$\pm$3.0 \\
4 & 204.6$\pm$19.6 & 208.0$\pm$20.4 & -3.4$\pm$3.6 & 4.1$\pm$2.8 \\
5 & 213.2$\pm$14.4 & 215.3$\pm$15.2 & -2.1$\pm$2.2 & 2.7$\pm$1.4 \\
6 & 201.2$\pm$21.4 & 200.8$\pm$24.1 & 0.3$\pm$4.8 & 3.8$\pm$3.0 \\
7 & 206.5$\pm$19.7 & 205.6$\pm$22.7 & 0.8$\pm$4.3 & 3.4$\pm$2.7 \\
8 & 212.8$\pm$16.7 & 214.4$\pm$18.4 & -1.6$\pm$2.9 & 2.9$\pm$1.7 \\
9 & 204.4$\pm$19.2 & 204.9$\pm$20.7 & -0.5$\pm$3.4 & 2.8$\pm$2.0 \\
10 & 204.6$\pm$21.6 & 208.4$\pm$21.7 & -3.8$\pm$2.9 & 4.2$\pm$2.3 \\
11 & 216.4$\pm$12.5 & 229.4$\pm$4.6 & -13.0$\pm$7.9 & 13.0$\pm$7.9 \\
12 & 210.0$\pm$18.0 & 210.3$\pm$21.2 & -0.3$\pm$4.8 & 3.0$\pm$3.7 \\
13 & 212.7$\pm$12.8 & 213.1$\pm$15.2 & -0.4$\pm$3.1 & 2.7$\pm$1.6 \\
14 & 206.5$\pm$17.4 & 210.9$\pm$16.8 & -4.5$\pm$2.7 & 4.7$\pm$2.3 \\
15 & 209.8$\pm$17.8 & 210.1$\pm$20.1 & -0.3$\pm$3.5 & 2.7$\pm$2.3 \\
16 & 201.3$\pm$21.7 & 205.2$\pm$22.3 & -4.0$\pm$4.1 & 4.8$\pm$3.0 \\
17 & 205.3$\pm$20.0 & 204.4$\pm$23.9 & 0.9$\pm$4.6 & 3.6$\pm$3.0 \\
18 & 198.7$\pm$14.3 & 193.1$\pm$17.2 & 5.6$\pm$7.3 & 7.0$\pm$6.0 \\
19 & 207.2$\pm$19.3 & 210.4$\pm$20.0 & -3.2$\pm$2.4 & 3.6$\pm$1.7 \\
20 & 209.8$\pm$16.0 & 209.8$\pm$18.0 & 0.0$\pm$4.2 & 3.2$\pm$2.7 \\
21 & 216.0$\pm$13.9 & 219.0$\pm$14.6 & -3.0$\pm$2.4 & 3.3$\pm$1.9 \\
22 & 203.3$\pm$20.2 & 203.1$\pm$23.1 & 0.3$\pm$4.0 & 3.1$\pm$2.4 \\
23 & 214.4$\pm$13.1 & 213.9$\pm$14.9 & 0.5$\pm$2.6 & 2.2$\pm$1.5 \\
24 & 212.1$\pm$14.5 & 211.2$\pm$17.5 & 0.9$\pm$3.9 & 3.3$\pm$2.3 \\
25 & 208.0$\pm$16.3 & 208.7$\pm$18.3 & -0.6$\pm$4.8 & 3.7$\pm$3.2 \\
26 & 203.0$\pm$20.0 & 205.9$\pm$21.4 & -2.9$\pm$3.5 & 3.9$\pm$2.4 \\
27 & 203.7$\pm$18.3 & 205.2$\pm$20.1 & -1.5$\pm$4.2 & 3.7$\pm$2.5 \\
28 & 208.7$\pm$17.0 & 204.2$\pm$23.1 & 4.5$\pm$7.0 & 5.8$\pm$6.0 \\
29 & 212.1$\pm$15.5 & 212.9$\pm$17.5 & -0.8$\pm$3.3 & 2.9$\pm$1.9 \\
30 & 202.0$\pm$22.0 & 204.5$\pm$23.2 & -2.5$\pm$2.4 & 3.1$\pm$1.6 \\
31 & 215.3$\pm$13.2 & 216.1$\pm$15.2 & -0.8$\pm$2.6 & 2.4$\pm$1.4 \\
32 & 209.8$\pm$15.3 & 215.3$\pm$14.0 & -5.5$\pm$2.9 & 5.5$\pm$2.8 \\
33 & 207.5$\pm$20.4 & 211.8$\pm$19.8 & -4.3$\pm$2.5 & 4.3$\pm$2.4 \\
34 & 207.7$\pm$19.8 & 207.5$\pm$21.9 & 0.3$\pm$4.6 & 3.3$\pm$3.2 \\
\midrule
Mean & 207.3$\pm$6.1 & 208.4$\pm$9.2 & -1.0$\pm$4.1 & 4.3$\pm$2.8 \\
Median & 207.6 & 209.2 & -0.7 & 3.6 \\
Q1     & 203.9 & 205.0 & -3.2 & 3.0 \\
Q3     & 212.1 & 213.0 &  0.3 & 4.3 \\
\bottomrule
\end{NiceTabular}
\end{minipage}
\hfill
\begin{minipage}[t]{0.32\textwidth}
\centering
\textbf{Blue Channel}\\[2pt]
\begin{NiceTabular}{r c c c c}
\toprule
WSI & Gen. & GT & Diff. & $|$Diff.$|$ \\
\midrule
1 & 194.8$\pm$15.2 & 180.3$\pm$24.5 & 14.6$\pm$15.4 & 15.4$\pm$14.5 \\
2 & 206.8$\pm$19.1 & 204.6$\pm$22.7 & 2.2$\pm$5.8 & 4.5$\pm$4.3 \\
3 & 217.6$\pm$17.5 & 221.1$\pm$17.6 & -3.5$\pm$2.3 & 3.5$\pm$2.3 \\
4 & 209.3$\pm$16.7 & 211.2$\pm$18.6 & -1.9$\pm$3.7 & 3.4$\pm$2.4 \\
5 & 217.4$\pm$11.9 & 219.4$\pm$12.7 & -2.0$\pm$2.2 & 2.6$\pm$1.4 \\
6 & 207.6$\pm$17.6 & 207.1$\pm$20.5 & 0.5$\pm$4.7 & 3.6$\pm$3.0 \\
7 & 211.8$\pm$16.0 & 210.4$\pm$19.4 & 1.4$\pm$4.4 & 3.7$\pm$2.7 \\
8 & 216.9$\pm$14.1 & 218.3$\pm$16.0 & -1.4$\pm$2.6 & 2.6$\pm$1.4 \\
9 & 210.4$\pm$15.7 & 210.6$\pm$17.4 & -0.2$\pm$3.1 & 2.6$\pm$1.8 \\
10 & 210.1$\pm$17.9 & 213.6$\pm$18.2 & -3.5$\pm$2.7 & 3.9$\pm$2.0 \\
11 & 219.9$\pm$10.2 & 230.9$\pm$4.1 & -11.0$\pm$6.2 & 11.0$\pm$6.2 \\
12 & 214.1$\pm$15.3 & 214.7$\pm$18.3 & -0.6$\pm$4.9 & 3.2$\pm$3.8 \\
13 & 217.5$\pm$10.0 & 218.0$\pm$12.2 & -0.5$\pm$2.6 & 2.3$\pm$1.3 \\
14 & 211.9$\pm$14.0 & 215.6$\pm$14.2 & -3.7$\pm$2.7 & 4.1$\pm$2.0 \\
15 & 214.3$\pm$14.9 & 213.4$\pm$18.2 & 1.0$\pm$4.3 & 3.1$\pm$3.1 \\
16 & 206.7$\pm$18.5 & 208.1$\pm$20.8 & -1.4$\pm$4.8 & 4.1$\pm$3.0 \\
17 & 211.3$\pm$16.1 & 210.8$\pm$19.5 & 0.4$\pm$4.1 & 3.2$\pm$2.5 \\
18 & 205.4$\pm$11.5 & 199.4$\pm$15.1 & 6.0$\pm$7.5 & 7.3$\pm$6.3 \\
19 & 212.2$\pm$16.1 & 214.2$\pm$17.6 & -2.0$\pm$2.8 & 2.9$\pm$1.8 \\
20 & 213.2$\pm$14.5 & 212.1$\pm$17.0 & 1.1$\pm$4.7 & 3.4$\pm$3.4 \\
21 & 218.2$\pm$13.0 & 221.4$\pm$13.3 & -3.2$\pm$2.2 & 3.5$\pm$1.8 \\
22 & 208.6$\pm$17.2 & 208.8$\pm$19.7 & -0.2$\pm$3.6 & 2.8$\pm$2.3 \\
23 & 218.4$\pm$10.7 & 218.0$\pm$12.6 & 0.4$\pm$2.5 & 2.1$\pm$1.3 \\
24 & 215.3$\pm$12.9 & 214.4$\pm$15.8 & 0.9$\pm$3.9 & 3.2$\pm$2.4 \\
25 & 212.3$\pm$13.7 & 214.8$\pm$14.7 & -2.5$\pm$4.3 & 4.0$\pm$3.0 \\
26 & 207.7$\pm$17.0 & 210.2$\pm$18.9 & -2.5$\pm$3.5 & 3.7$\pm$2.2 \\
27 & 209.0$\pm$15.4 & 210.3$\pm$17.2 & -1.3$\pm$4.0 & 3.5$\pm$2.4 \\
28 & 213.4$\pm$14.4 & 209.1$\pm$21.2 & 4.3$\pm$7.7 & 5.7$\pm$6.7 \\
29 & 216.1$\pm$13.0 & 217.5$\pm$14.5 & -1.5$\pm$2.9 & 2.8$\pm$1.8 \\
30 & 206.8$\pm$19.0 & 209.1$\pm$20.4 & -2.3$\pm$2.7 & 3.1$\pm$1.5 \\
31 & 219.6$\pm$10.2 & 220.7$\pm$12.0 & -1.1$\pm$2.2 & 2.2$\pm$1.2 \\
32 & 213.9$\pm$12.8 & 217.4$\pm$12.8 & -3.6$\pm$2.1 & 3.7$\pm$1.9 \\
33 & 212.4$\pm$17.0 & 216.2$\pm$16.7 & -3.7$\pm$2.2 & 3.9$\pm$2.0 \\
34 & 212.4$\pm$16.5 & 212.3$\pm$18.9 & 0.1$\pm$4.3 & 3.0$\pm$3.0 \\
\midrule
Mean & 212.2$\pm$5.1 & 212.8$\pm$8.1 & -0.6$\pm$3.9 & 4.0$\pm$2.6 \\
Median & 212.4 & 213.5 & -1.2 & 3.4 \\
Q1     & 209.1 & 210.2 & -2.4 & 2.9 \\
Q3     & 215.9 & 217.5 &  0.5 & 3.9 \\
\bottomrule
\end{NiceTabular}
\end{minipage}
\par\smallskip
{\justifying\scriptsize
Mean channel values ($\pm$ SD) for generated (Gen.) and ground truth (GT) patches by origin WSI. Diff.\ is the signed mean difference; $|$Diff.$|$ is the mean absolute difference. All values on a 0--255 scale. WSIs 1 and 18 show large RGB deviations, consistent with less favorable LPIPS and SSIM scores in Table~\ref{table:test-metrics}.\par}
\label{table:rgb-metrics}
\end{table*}
\end{document}